\pdfoutput=1
\documentclass[10pt,twocolumn,letterpaper]{article}

\usepackage[pagenumbers]{cvpr} % To force page numbers, e.g. for an arXiv version

\usepackage[export]{adjustbox}
\usepackage{array}
\usepackage{multirow}
\usepackage{arydshln}
\usepackage{amssymb}
\usepackage{MnSymbol}
\usepackage{pifont}
\newcommand{\cmark}{\ding{51}}%
\newcommand{\xmark}{\ding{55}}%
\usepackage{booktabs}
\usepackage[dvipsnames]{xcolor}
\usepackage[nolist]{acronym}
\usepackage{csquotes}
\usepackage{tikz}

\begin{acronym}
	\acro{HD}{high-definition}
\end{acronym}

\DeclareMathAlphabet\mathbfcal{OMS}{cmsy}{b}{n}

\newcommand{\pointembedding}{Q_\mathrm{pt}}
\newcommand{\positionalembedding}{Q_\mathrm{PE}}
\newcommand{\pointpriorembedding}{E_\mathrm{pt}}
\newcommand{\positionalpriorembedding}{E_\mathrm{PE}}
\newcommand{\referencepoint}{P_\mathrm{ref}}
\newcommand{\AV}{Argoverse~2\xspace}
\newcommand{\data}{\mathcal{D}}
\newcommand{\map}{\mathcal{M}}

\newcommand{\gtmap}{\map_{\mathrm{GT}}}
\newcommand{\scenario}{\mathcal{S}}
\newcommand{\scenarios}{\mathbfcal{S}}
\newcommand{\priorgenerator}{P}
\newcommand{\expert}{\emph{Expert}\xspace}
\newcommand{\experts}{\emph{Experts}\xspace}
\newcommand{\generalist}{\emph{Generalist}\xspace}
\newcommand{\mapc}{mAP$^{\mathbfcal{C}}$\xspace}

\newcommand{\circled}[1]{\raisebox{.5pt}{\textcircled{\raisebox{-.9pt} {#1}}}}

\definecolor{cvprblue}{rgb}{0.21,0.49,0.74}
\usepackage[pagebackref,breaklinks,colorlinks,allcolors=cvprblue]{hyperref}

\def\paperID{14454} % *** Enter the Paper ID here
\def\confName{ICCV}
\def\confYear{2025}

\title{M3TR: A Generalist Model for Real-World HD Map Completion}
\author{
Fabian Immel$^{1}$ \quad Richard Fehler$^{1}$ \quad Frank Bieder$^{1}$ \quad Jan-Hendrik Pauls$^{2}$ \quad Christoph Stiller$^{2}$\\
$^{1}$FZI Research Center for Information Technology \quad $^{2}$Karlsruhe Institute of Technology\\
{\tt\small \{immel, fehler, bieder\}@fzi.de} \quad {\tt\small \{jan-hendrik.pauls, stiller\}@kit.edu}
}

\begin{document}
\maketitle
\begin{abstract}

    Autonomous vehicles rely on HD maps for their operation, but offline HD maps eventually become outdated.
    For this reason, online HD map construction methods use live sensor data to infer map information instead.
    Research on real map changes shows that oftentimes entire parts of an HD map remain unchanged and can be used as a prior.
    We therefore introduce M3TR (Multi-Masking Map Transformer), a generalist approach for HD map completion both with and without offline HD map priors.
    As a necessary foundation, we address shortcomings in ground truth labels for Argoverse~2 and nuScenes and propose the first comprehensive benchmark for HD map completion.
    Unlike existing models that specialize in a single kind of map change, which is unrealistic for deployment, our Generalist model handles all kinds of changes, matching the effectiveness of Expert models.
    With our map masking as augmentation regime, we can even achieve a $+1.4$ mAP improvement without a prior.
    Finally, by fully utilizing prior HD map elements and optimizing query designs, M3TR outperforms existing methods by $+4.3$ mAP while being the first real-world deployable model for offline HD map priors.
    \href{https://github.com/immel-f/m3tr}{\textnormal{\texttt{https://github.com/immel-f/m3tr}}}
    
\end{abstract}
\section{Introduction}
\label{sec:intro}

In order to drive safely, autonomous vehicles need to understand the geometry and topology of the roads as well as the traffic rules that apply to them.
Current systems employ detailed semantic \ac{HD} maps that provide this rich knowledge, but are primarily created using offline SLAM approaches.
However, maintaining such offline \ac{HD} maps to account for changes in no time is infeasible.
Therefore, recent advances in computer vision aim to perceive \ac{HD} map information with onboard sensors~\cite{HDMapNet,vectormapnet, maptr, maptrv2, chen2024maptracker, Zhou_2024_CVPR}.

This task of online vectorized \textit{HD map construction} uses sensor data, \eg from cameras or LiDAR sensors, to detect vectorized map elements (lane markings, road borders, \etc) with their semantic meaning. 
Compared to offline HD planning maps however, the output of online HD map construction models still lacks a large amount of information.

Recent research~\cite{av2_trust_but_verify} showed that maps only gradually become outdated and that oftentimes some parts of vectorized (offline) map information is still up-to-date and could be used as prior. 
Concretely, only specific, often semantically coherent elements are invalidated, while leaving the rest unchanged. 
This leads to the situation where a map perception model needs to fill in the invalid parts using the remaining HD map and online sensor information, a task which we refer to as \emph{HD map completion}.

Existing work that incorporates prior information falls short for three main reasons:
While detection transformer queries are used to provide vectorized priors to the model \cite{sun2024mapex}, they fail to fully utilize all prior map information.
Furthermore, current approaches lack a clear task definition and evaluation metric that can differentiate prior map elements and those that need to be perceived online. 
Finally and most importantly, previous models specialize on a single kind of map prior that is assumed to be known in advance.
Since any part of an offline HD map could change, this is an unrealistic expectation for real-world deployment.

\begin{figure*}
    %\hspace*{-1cm}
    \centering
    \includegraphics[trim={0.45cm 0cm 0cm 0cm},clip,width=\linewidth]{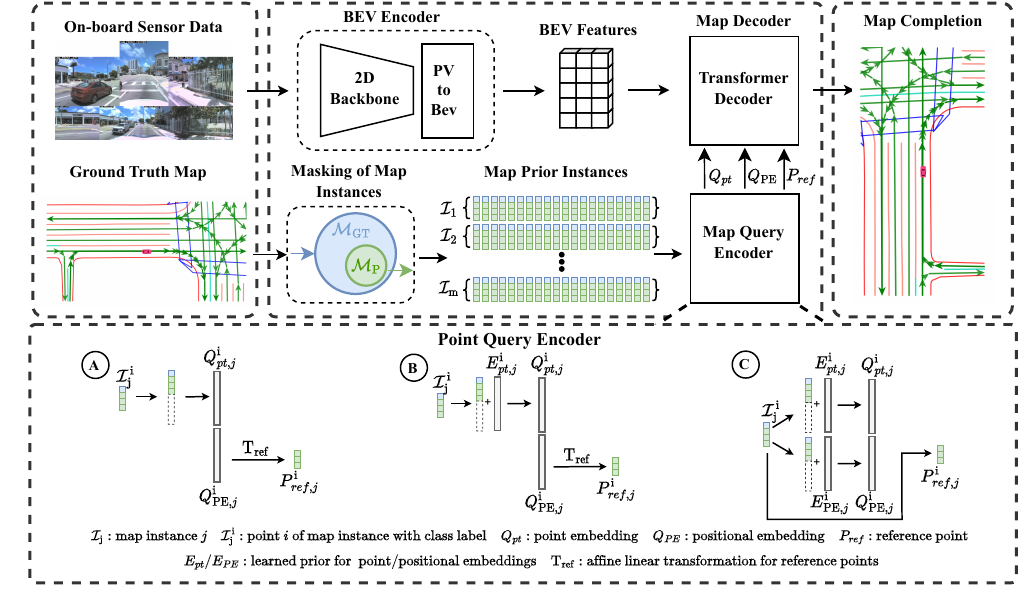}
    \caption{Overview of the model architecture of M3TR and the investigated point query encoder designs. For our evaluated task of HD map completion, we mask out instances from the ground truth map~$\mathcal{M_{\mathrm{GT}}}$ to create a map prior~$\mathcal{M_{\mathrm{P}}}$. Using $\mathcal{M_{\mathrm{P}}}$,  we try to reconstruct~$\mathcal{M_{\mathrm{GT}}}$. The map prior instances are supplied to the model as queries, influenced by the shown point query encoder and the detection query set design which is further illustrated in \cref{fig:o2m_queries}.
    }
    \label{fig:overview_flowchart}
\end{figure*}

\subsection*{Contributions}

To address these points, we present M3TR (Multi-Masking Map Transformer), a generalist HD map completion model with the following contributions:
\begin{itemize}
    \item A new HD map completion benchmark for models with prior offline HD map information. This includes %new task definitions with 
    semantically richer labels and the first metric that explicitly focuses on the performance for elements \emph{without} a prior.
    \item We propose a novel query design to incorporate map priors on a point query and query set level that considerably improves detection performance on the Argoverse~2 and nuScenes datasets by up to $+4.3$ mAP.
    \item We introduce a novel training regime which yields a single model that can make use of any HD map prior.  
    This \emph{Generalist} model achieves performance on par with specialized models without needing to know which kind of map information is available, even improving performance without a prior by up to 1.4 mAP.
\end{itemize}

\section {Related Work}
\label{sec:related_work}

Related work can be grouped into two main categories: Common online HD map construction methods without priors and methods that use prior vectorized map information. 

\subsection{Online HD Map Construction without Priors}
\label{subsec:online_hd_map_construction}

Detection transformer (DETR)~\cite{DETR_2020ECCV} based architectures can be used to provide vectorized map element detections, handling HD map polyline and polygon elements in their original sparse representation. 
To detect consistent map elements in the surrounding scene, a bird's eye view (BEV) feature grid, representing a fixed environment area, is generated by transforming 2D image features with methods proposed in general 3D object detection and BEV segmentation~\cite{philion2020liftsplatshoot,li2022bevformer,chen2022efficientGKT}.
MapTR~\cite{maptr,maptrv2} implements fast detection for complete map elements by modifying the original object queries of the transformer decoder to represent polylines and polygons with a fixed number of points.
This enables fast parallel transformer decoding in contrast to early autoregressive approaches like VectorMapNet~\cite{vectormapnet}.  

Recent contributions in the online HD map construction task show two significant improvements to the MapTR and MapTRv2 baselines, concentrating on query design and formulation.
The first set \cite{Zhou_2024_CVPR,choi2024mask2map} improves single-shot detection performance by utilizing complete map element shapes and masks in the detection query representation. 

The second set \cite{Yuan_2024_streammapnet,chen2024maptracker} extends the single-shot detection task to the temporal and spatial context of past time steps. 

\subsection{Online HD Map Construction with Priors}
\label{subsec:map_prior}
Methods discussed in the previous section take only sensor data into account. 
In real-world autonomous systems, maps ranging from navigation maps to HD maps are used for at least routing, extending to motion prediction, path planning, and other driving tasks. 
Since this map becomes outdated piece by piece, online HD map construction that utilizes still up-to-date parts of maps as an optional prior is an attractive solution from an application perspective. 

MapEX~\cite{sun2024mapex} was among the first to propose a detection query design allowing for both existing map element transformer queries as prior and regular learned transformer queries used for detecting unknown map elements.
We use it as a baseline, but improve not only its evaluation scheme, but also the query design and model capabilities.
          
PriorDrive~\cite{zeng2024unifiedvectorpriorencoding} proposes a HD map construction framework which integrates either SD navigation maps, incomplete HD maps or online constructed HD maps from previous drives at the same location.
SMERF~\cite{luo2023augmentingSMERF} incorporates a SD map prior by first encoding SD map elements with a transformer encoder and fusing them with the BEV feature grid via cross attention, showing improvements on the OpenLaneV2~\cite{openlaneV2} dataset detection and topology metrics. 
The approach of \cite{sun2024mapex} to incorporate existing map prior with varying degradation levels was extended by \cite{Bateman2024CVPR} and consecutively~\cite{wild2024exelmap} to use heavily modified map prior inputs to simulate outdated and incorrect map priors.
This expands the training task to map verification, change detection, and map update, showing a significant sim-to-real gap on real public \cite{av2_trust_but_verify} or proprietary 
\cite{Bateman2024CVPR} data.

As mentioned in \cref{sec:intro}, previous works show a number of shortcomings, which will be discussed in the next section along with our improvements.

\section{The HD Map Completion Benchmark}
In this section we describe the novel HD map completion benchmark.
\cref{subsec:ground_truth_generation} discusses our improved ground truth while \cref{subsec:hd_map_completion} presents the proposed HD map completion task.
\label{sec:method}

\subsection{Improved Ground Truth Maps for Real World Autonomous Driving}
\label{subsec:ground_truth_generation}

\begin{table}[b]%[b]
\centering
\caption{Features of labels on Argoverse 2 used in various state of the art approaches and in our proposed ground truth. }
\resizebox{\columnwidth}{!}{%
\begin{tabular}{l wc{1.0cm} wc{1.0cm} wc{1.0cm} wc{1.0cm} wc{1.0cm}} 
\multirow{2}{*}{\textbf{Method}} & \multirow{2}{*}{\shortstack[l]{Divider \\ Types}} & \multirow{2}{*}{\shortstack[l]{Lane \\ Centerl.}} &  \multirow{2}{*}{\shortstack[l]{3D Ins-\\ tances}} & \multirow{2}{*}{\shortstack[l]{Fixed GT \\Artifacts}} & \multirow{2}{*}{\shortstack[l]{Geo. \\ Split}} \\ 
\\
\toprule
VectorMapNet \cite{vectormapnet} & - & - & - & - & - \\
MapTRv2 \cite{maptrv2, lanegap} & - & $\checkmark$ & $\checkmark$ & - & - \\  
StreamMapNet \cite{Yuan_2024_streammapnet} & - & - & - & - & $\checkmark$ \\
MapTracker \cite{chen2024maptracker} & - & - & - & $\checkmark$ & $\checkmark$ \\
MapEX \cite{sun2024mapex} & - & - & - & - & - \\
PriorDrive \cite{zeng2024unifiedvectorpriorencoding} & - & - & - & - & - \\
\textbf{M3TR (Ours)} & $\checkmark$ & $\checkmark$ & $\checkmark$ & $\checkmark$ & $\checkmark$ \\ 
\bottomrule
\end{tabular}
}
\label{tab:label_comparison}
\vspace{-1mm}
\end{table}

Most recent HD map construction models are trained using labels that have largely been unchanged since VectorMapNet~\cite{vectormapnet} despite having major shortcomings. 

The labels lack information that is necessary for autonomous driving, issues in the label generation algorithms introduce errors into the ground truth instances and a geographic overlap leads to leakage between training and evaluation data.
Fixes to these issues have been proposed in different works~\cite{maptrv2, lanegap, chen2024maptracker, Lilja2024CVPR}, however they are scattered and not united in one single ground truth set. 

Hence, as foundation for the proposed HD map completion benchmark, we combine these improvements into one label set together with a novel separation of dashed and solid dividers.
A comparison with previously used labels is listed in \cref{tab:label_comparison}.
For a detailed description with qualitative examples we refer to the supplementary material.

\subsection{The HD Map Completion Task}
\label{subsec:hd_map_completion}

Our focus in this work is on using offline HD maps as priors that became outdated and thus partially invalid.
This idea is based on the largest public dataset of real map changes, Trust but Verify \cite{av2_trust_but_verify}.
When applying existing work to outdated offline HD maps, three open issues arise:
Map changes in public datasets are not labeled on a point level and rare, requiring map priors to be derived synthetically~\cite{av2_trust_but_verify}.
Unfortunately, the map change generation schemes of previous approaches~\cite{sun2024mapex, zeng2024unifiedvectorpriorencoding, wild2024exelmap} follow assumptions that are not applicable for outdated offline HD maps.
Real changes do not occur at random, but rather follow a local pattern with semantic correlation~\cite{Bateman2024CVPR}, that only affects specific elements and leaves most elements unchanged.
This was also observed in \cite{av2_trust_but_verify}, where changes remove, modify or add semantically coherent elements rather than randomly drop/add elements or apply noise.

\begin{table}[b]%[htb]
\centering
\caption{Systematic map prior scenarios $\scenarios$ defined in this work.}
\begin{tabular}{ll} 
\textbf{Name} & \textbf{Description} \\
\toprule
$\scenario_{\overline{\mathrm{EL}}}$ & Ego lane is masked out.\\
$\scenario_{\overline{\mathrm{ER}}}$ & Ego road is masked out.\\
$\scenario_{\mathrm{BD}}$ & Only road boundaries are provided as prior.\\
$\scenario_{\mathrm{CL}}$ & Only lane center lines are provided as prior.\\
$\scenario_{\varnothing}$ & No map prior.\\
\bottomrule
\end{tabular}
\label{tab:prior_scenarios}
\vspace{-1mm}
\end{table}

To better align our synthetic priors with real priors, we define adapted map prior scenarios $\scenario_p = (\map_p, \data)$ consisting of a map prior $\map_p$ and sensor data $\data$.
Map priors $\map_p$ are derived from the complete ground truth map $\map_{\mathrm{GT}}$ using the scenario specific prior generator $\priorgenerator_p$ which masks out or selects only specific map elements:
\begin{equation}
    \map_p = \priorgenerator_p(\gtmap).
\end{equation}
The task of the model is to reconstruct the complete map from the given partial prior and the sensor information.

\begin{figure}
    \centering
    \begin{subfigure}[t]{0.48\linewidth}
        \centering
        \includegraphics[trim={13.0cm 0.0cm 0.0cm 0.0cm},clip,width=\linewidth]{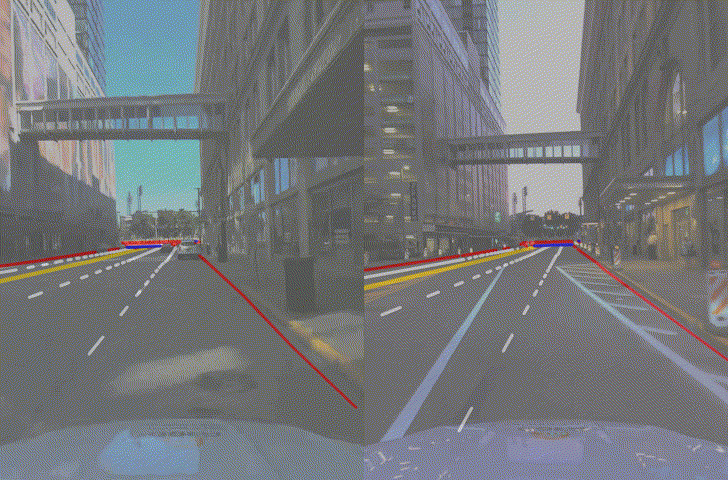} 
        \caption{Ex. for $\scenario_{\overline{\mathrm{EL}}}$: Own lane blocked.}
        \label{fig:av2_tbv_samples_sub1}
    \end{subfigure}
    \hfill
    \begin{subfigure}[t]{0.48\linewidth}
        \centering
        % trim={<left> <lower> <right> <upper>}
    \includegraphics[trim={13.0cm 0.0cm 0.0cm 0.0cm},clip,width=\linewidth]{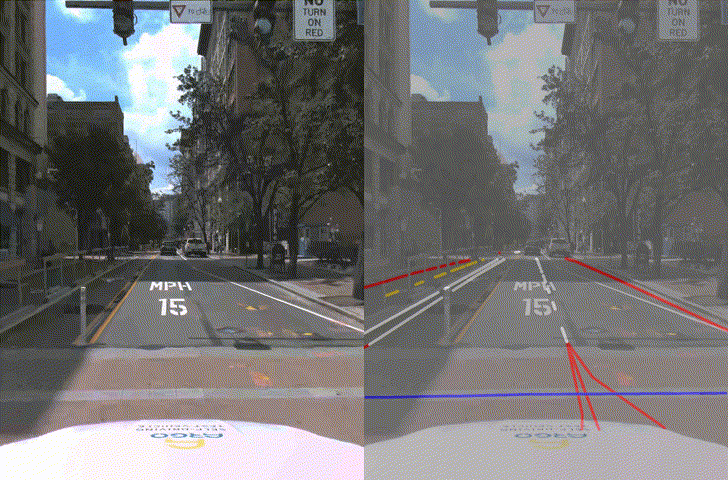} 
        \caption{Ex. for $\scenario_{\overline{\mathrm{ER}}}$: New bike lane.}
        \label{fig:av2_tbv_samples_sub2}
    \end{subfigure}
    \caption{Visualization of map changes from \cite{av2_trust_but_verify}, with the outdated map reprojected into the camera image.
    Real map changes can easily be translated into the proposed map prior scenarios.}
    \label{fig:av2_tbv_samples}
\end{figure}

MapEX~\cite{sun2024mapex} has already begun moving in this direction by including $\scenario_{\mathrm{BD}}$ as a scenario, however the other scenarios include modifications like point-level noise that are not applicable for outdated offline HD maps, but only for maps perceived in previous time steps.

We show in \cref{sec:experiments} that the semantic class of map prior has a strong influence on the model performance and therefore propose to separate map prior scenarios semantically. 
This enables a systematic investigation to guide future efforts in data collection or map maintenance.
The prior scenarios are listed in \cref{tab:prior_scenarios} and visualized in \cref{fig:expert_generalist}.
\cref{fig:av2_tbv_samples} shows that real map changes~\cite{av2_trust_but_verify} can easily be categorized into the proposed scenarios.
\cref{fig:av2_tbv_samples_sub1} has the own lane become blocked, resulting in invalidated elements akin to $\scenario_{\overline{\mathrm{EL}}}$. In \cref{fig:av2_tbv_samples_sub2}, a bike lane is added that causes the ego road to become invalid, similar to $\scenario_{\overline{\mathrm{ER}}}$.

The scenarios assume that it is known beforehand which elements are no longer valid, a task for which separate proposed solutions exist \cite{wild2024exelmap, map_ver_pauls}. 
In turn however, this also brings a large benefit: 
These map prior scenarios avoid a cause of the significant sim-to-real gap already noted for artificial map changes~\cite{Bateman2024CVPR}, namely the fact that most synthetic changes are not logically consistent with the sensor data.
This is because the reconstruction task is indifferent to whether elements are masked synthetically or if elements become masked due to real map changes, as both mask semantically coherent elements.

\subsection{A Prior-Aware HD Map Completion Metric}

To measure map completion performance, we need to solve an issue already pointed out by~\cite{wild2024exelmap}: current evaluation metrics do not differentiate between map elements that are available as prior and those that need to be perceived online~\cite{sun2024mapex, zeng2024unifiedvectorpriorencoding, Bateman2024CVPR}.
However, transformer models quickly learn to pass through prior elements almost identically and, if known as prior, any downstream application would prefer the map prior over the corresponding, possibly noisy prediction.
Hence, we propose to focus on exactly those map elements which are unknown to the model at inference time.

The standard evaluation metric for methods with vectorized output~\cite{maptrv2, chen2024maptracker, vectormapnet} is the mean average precision (mAP), using the Chamfer distance with thresholds of $\tau \in \{0.5 \mathrm{m}, 1.0 \mathrm{m}, 1.5 \mathrm{m} \}$. 
The mAP is averaged across the average precision (AP) of the individual label classes: dashed dividers, solid dividers, road boundaries, lane centerline paths and pedestrian crossings, with the class specific AP averaged across the Chamfer distance thresholds $\tau$.
Analogously, to evaluate completion performance, we define the mean average \emph{completion} precision, %\todo{(ganz schön sperrig)} 
$\textrm{mAP}^{\mathcal{C}}$, which uses not the entire map $\gtmap$, but only the map elements $\map_{\overline{p}} = \gtmap \mathbin{\backslash} \map_p$ which are missing in the specific scenario.

\section{A Deployable HD Map Completion Model}

This section describes the novel M3TR (Multi-Masking Map Transformer) model itself.
\cref{subsec:training_regime} presents our \generalist training regime, \cref{subsec:map_masking_augmentation} how to use map masking as augmentation, and \cref{subsection:query_design} the novel map prior query design.

\subsection{Query Design}
\label{subsection:query_design}
In recent work, queries of the detection transformer have emerged as the main way to supply the model with prior information~\cite{sun2024mapex, wild2024exelmap, zeng2024unifiedvectorpriorencoding, Bateman2024CVPR, luo2023augmenting, chen2024maptracker}. 
% How exactly these queries are composed and where they are inserted is often neglected and not described in detail. 
BEV detection transformer queries consist of different sub-elements and map prior queries can be composed in many different ways and inserted at multiple points, making the available option space quite large. 
We explore that option space to incorporate prior map knowledge on two architectural levels, the \textit{point query design} and the \textit{query set design}, and use MapEX~\cite{sun2024mapex} as our baseline approach.

\subsubsection*{Point Query Design}

For each map element that is to be predicted, a fixed set of points is used as map decoder queries.
Each point query consists of two vectors which are concatenated: the point embedding $\pointembedding$ and the positional embedding $\positionalembedding$.

Most HD map construction transformers use learned point embeddings since they assume no prior knowledge about map elements.
To improve upon this, we compare three approaches, \circled{A} -- \circled{C} depicted in \cref{fig:overview_flowchart}, on how to encode map prior information into the point queries, increasing performance with our novel approach \circled{C}.

While in \circled{A}, the baseline proposed in MapEX~\cite{sun2024mapex}, the zero-padded point information is directly used as point embedding $\pointembedding$, we propose to combine it with a two-part learned prior embedding $\pointpriorembedding$ in \circled{C}.
This makes use of the prior information, but provides a learnable degree of freedom for the model.

\circled{B}, a learned embedding design also explored in \cite{sun2024mapex}, differs from \circled{C} in the positional embedding $\positionalembedding$.
It is formed by either a sum of zero-padded point information for \circled{A} and \circled{B} or a learned prior embedding $\positionalpriorembedding$ for \circled{C}.

To each query also belongs a reference point on the BEV grid, $\referencepoint$, which guides the deformable cross-attention in the decoder.
In the previous designs \circled{A} and \circled{B}, it is generated from the positional embeddings with a linear projection.
To improve upon this, in \circled{C} we propose to directly define it based on the map prior point information.

\begin{figure}
    \centering
    % trim={<left> <lower> <right> <upper>}
    \includegraphics[trim={0.6cm 0.3cm 0.4cm 0},clip,width=0.99\linewidth]{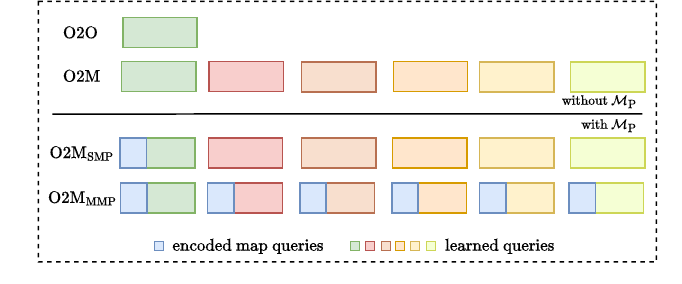}
    \caption{Visualization of different detection query set designs with and without map prior $\map_p$.
    The set of queries are matched to ground truth map elements in either a one-to-one ($\mathrm{O2O}$) or one-to-many ($\mathrm{O2M}$) fashion.
    Compared to the baseline $\mathrm{O2M_{SMP}}$ query set design for map priors, we propose a tiling $\mathrm{O2M_{MMP}}$ design.}
    \label{fig:o2m_queries}
\end{figure}

\begin{figure}
    % trim={<left> <lower> <right> <upper>}
    \includegraphics[trim={1.1cm 0.1cm 0.5cm 0},clip,width=0.99\linewidth]{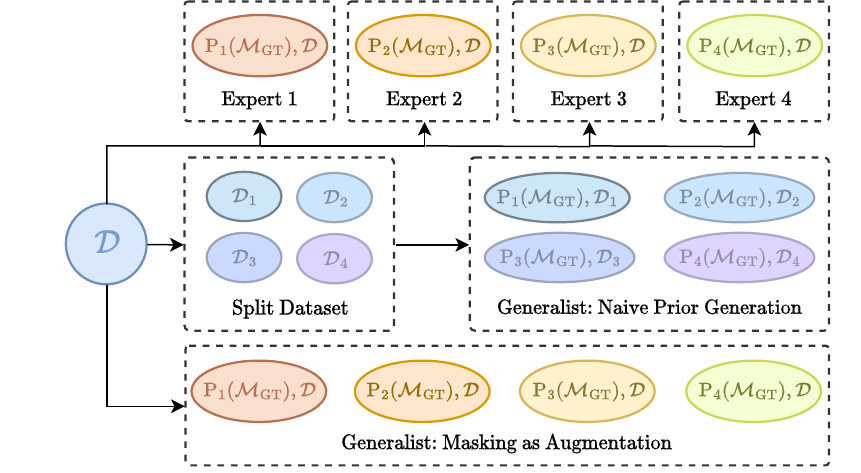}
    \caption{
    Visualization of the different training regimes for variable map priors investigated in this work.
    Compared to previous expert training regimes and a naive \generalist prior generation, our masking as augmentation regime leverages all available data for a \generalist model with improved performance.}
    \label{fig:training_regimes}
\end{figure}

\subsubsection*{Query Set Design}
MapTRv2~\cite{maptrv2} proposed one-to-many (O2M) matching, a source of significant performance gains compared to the original MapTR one-to-one (O2O) matching.
We explore two possible ways to adapt it to map prior information which are depicted in \cref{fig:o2m_queries}.

In the $\mathrm{O2M}_\mathrm{SMP}$ (\emph{Single Map Prior}) query set design only a single repetition of queries makes use of map prior queries while auxiliary queries are purely learned, just like in the original MapTRv2.
In contrast, the $\mathrm{O2M}_\mathrm{MMP}$ (\emph{Multiple Map Prior}) query set design includes map prior information in a tiling fashion, once for every repetition of the ground truth.
This allows the incorporation of map prior knowledge in the auxiliary queries as well.

We follow MapEX~\cite{sun2024mapex} for the loss, including the pre-attribution of map prior instances during the Hungarian assignment, which we extend to the tiled $\mathrm{O2M}_\mathrm{MMP}$ map prior queries.
Outside of instances related to map priors, the MapEX loss is equivalent to the loss of the MapTRv2 base architecture.
A more detailed description of the pre-attribution can be found in the supplementary material.

\subsection{Generalist and Expert Models}
\label{subsec:training_regime}

\begin{figure}
    \centering
    \begin{subfigure}[t]{0.85\linewidth}
        \centering
        \includegraphics[trim={0.8cm 0.8cm 0.8cm 0.7cm},clip,width=\linewidth]{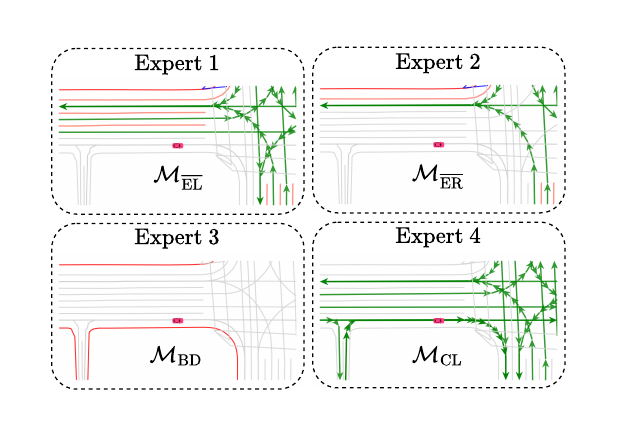} 
        \caption{One expert model per map prior scenario.}
        %\label{fig:sub1}
    \end{subfigure}
    \hfill
    \begin{subfigure}[t]{0.85\linewidth}
        \centering
        % trim={<left> <lower> <right> <upper>}
    \includegraphics[trim={0.95cm 1.0cm 0.4cm 0.9cm},clip,width=\linewidth]{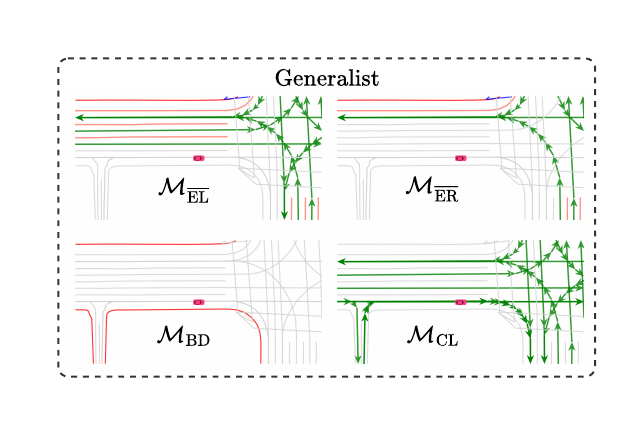} 
        \caption{One \generalist model for all map prior scenarios.}
        %\label{fig:sub2}
    \end{subfigure}
    \caption{Visualization of previous expert models vs. the \generalist model proposed in this work. The map prior scenarios $\scenario_p$ are listed in \cref{tab:prior_scenarios}.}
    \label{fig:expert_generalist}
\end{figure}

Previous works train one model for each map prior scenario $\scenario_p$ which is unrealistic for deployment in real autonomous systems.
All models would need to be readily available in GPU memory and the suitable model would need to be correctly selected by a not yet existing oracle that identifies the available prior category.
We refer to these models as \experts and instead propose a \generalist model that can exploit arbitrary parts of HD maps as a prior.
Instead of only one, the \generalist is trained on \emph{all} scenarios $\scenarios = \cup \, \scenario_p$.
As we show below, while needing no extra memory or compute, it is on par with specialized \experts.
We visualize the distinction in \cref{fig:expert_generalist}.

\subsection{Map Masking as Augmentation}
\label{subsec:map_masking_augmentation}

To train a \generalist with synthetically derived map priors, various training regimes are conceivable.
As depicted in \cref{fig:training_regimes}, to derive $n$ map prior scenarios, one could naively split the dataset $\data$ into $n$ equal disjoint smaller datasets $\data_i$ and use each part to derive one kind of prior scenario $\scenario_i$:
\begin{equation}
    \scenario_i= (\map_{p_i}, \data_i) = (\priorgenerator_{p_i}(\gtmap), \data_i).
\end{equation}

Instead, we propose to use the synthetic generation of prior scenario as augmentation for generic HD map construction.
This means that the entire dataset $\data$ is used to derive each map prior scenario $\scenario^\#_p$ and, hence, the augmented scenario set $\scenarios^\#$:
\begin{align}
    \scenario^\#_i &= (\map_p, \data) = (\priorgenerator_p(\gtmap), \data)\\
    \scenarios^\# &= \cup_p \scenario^\#_p.
\end{align}

This exploits the entire combinatorial variety of dataset diversity and map prior categories and leads to an $n$-fold increase in training data, promising greater generalization performance.

\section{Experiments}
\label{sec:experiments}

We conduct experiments on the \AV and nuScenes datasets to validate our method, with \AV as the main dataset. 
\cref{subsec:dataset_metric} elaborates on the choices of dataset and metric, \cref{subsec:implementation} on the implementation and \cref{subsec:performance} discusses the performance of M3TR in comparison with existing baselines.

\subsection{Dataset and Metric}
\label{subsec:dataset_metric}

% As mentioned above, we validate our method on the \AV~\cite{Argoverse2} and nuScenes datasets~\cite{nuscenes}, the standard public datasets for HD map construction.
Both \AV and nuScenes contain 1000 driving sequences, covering 17~km² and 5~km², respectively~\cite{Lilja2024CVPR}. 
Since nuScenes has only 40,000 samples compared to 158,000 for \AV, contrary to most existing work, we regard \AV as our primary dataset for evaluation.
As discussed in \cref{subsec:ground_truth_generation}, we use a novel kind of ground truth that resolves a number of problems compared to the labels used in~\cite{maptr, maptrv2, vectormapnet, Yuan_2024_streammapnet}.

As our metric, we use the mean average \emph{completion} precision $\textrm{mAP}^{\mathcal{C}}$ defined in \cref{subsec:hd_map_completion} to compare the methods.

%%%%%%%%%%%%%%%%%%%%%%%%%%%%%%%%%%%%%%%%%%%%%%%%%%%%%%%%%%%%%%%%%%%%%%%%%%%%%%%%%%%%%%%%%%%%

\begin{table*}
\centering
\caption{Comparison of methods over map prior scenarios on the Argoverse 2 data set, with the geographical split from~\cite{Lilja2024CVPR}.
Only elements not in the map prior are evaluated.
$\map_{\overline{\mathrm{EL}}}$: Ego lane is masked.
$\map_{\overline{\mathrm{ER}}}$: Ego road is masked.
$\map_{\mathrm{BD}}$: Only road boundaries as prior.
$\map_{\mathrm{CL}}$: Only centerlines as prior.
$\map_{\varnothing}$: No map prior.
$\mathbfcal{O}(|\map_{\mathrm{P}}|)$ indicates how the deployment effort scales with the number of map prior types.
The last column indicates whether the method could handle variable priors without a not yet existing scenario to expert assignment oracle.
*: Re-implemented by the authors, as code was not publicly available at the time of publication.
\textdagger: For no map prior both expert methods are equivalent to MapTRv2, as changes in the base architecture are only made regarding map priors.}
\setlength\dashlinedash{1.2pt}
\setlength\dashlinegap{2.0pt}
\setlength\arrayrulewidth{0.3pt}
\resizebox{0.95\linewidth}{!}{%
\begin{tabular}{l l wr{1.0cm} wr{1.0cm} wr{1.0cm} wr{1.0cm} wr{1.0cm} wr{1.0cm} wr{1.0cm} wc{1.2cm} wc{1.2cm}}
\toprule
\multicolumn{2}{c}{\small{\textit{Dataset: Argoverse 2}}} & \multicolumn{7}{c}{\small{\textit{$\textrm{AP}^{\mathcal{C}}$ = AP for Masked Elements only}}} & \multirow{2}{*}{\small{$\mathbfcal{O}(|\map_{\mathrm{P}}|)$}} & \multirow{2}{*}{\small{\textbf{\shortstack[c]{Var. Prior \\ w/o Oracle}}}} \\
\textbf{Method} & \textbf{Map Prior} & $\textbf{AP}^{\mathbfcal{C}}_{\textbf{dsh}}$ & $\textbf{AP}^{\mathbfcal{C}}_{\textbf{sol}}$ & $\textbf{AP}^{\mathbfcal{C}}_{\textbf{bou}}$ & $\textbf{AP}^{\mathbfcal{C}}_{\textbf{cen}}$ & $\textbf{AP}^{\mathbfcal{C}}_{\textbf{ped}}$ & $\textbf{mAP}^{\mathbfcal{C}}$ & \textbf{vs.~\cite{sun2024mapex}} & & \\ 
\vspace{-0.35cm} \\ \toprule 
MapTRv2~\cite{maptrv2} & $\map_{\varnothing}$ & 37.9 & 55.0 & 49.7 & 48.2 & 41.7 & \textbf{46.5} & \textcolor{Gray}{\textsuperscript{\textdagger}\textbf{+0.0}} & - & - \\ \midrule
\multirow{6}{*}{\shortstack[l]{MapEX*~\cite{sun2024mapex} \\ Models}} & $\map_{\overline{\mathrm{EL}}}$ & 45.3 & 64.5 & 53.4 & 52.8 & 44.9 & \textbf{52.2} & - & \multirow{7}{*}{$\mathcal{O}(\mathrm{n})$} & \multirow{7}{*}{\textcolor{Red}{\textbf{\xmark}}} \\
& $\map_{\overline{\mathrm{ER}}}$ & 41.5 & 62.4 & 54.9 & 55.3 & 45.5 & \textbf{51.9} & - &  & \\
& $\map_{\mathrm{BD}}$ & 37.7 & 56.0 & - & 50.6 & 44.5 & \textbf{47.2} & - &  & \\
& $\map_{\mathrm{CL}}$ & 43.2 & 61.8 & 58.1 & - & 42.8 & \textbf{51.5} & - &  & \\
& $\map_{\varnothing}$ & \textsuperscript{\textdagger}37.9 & \textsuperscript{\textdagger}55.0 & \textsuperscript{\textdagger}49.7 & \textsuperscript{\textdagger}48.2 & \textsuperscript{\textdagger}41.7 & \textsuperscript{\textdagger}\textbf{46.5} & - &  & \\ \cdashline{2-9}
\vspace{-0.35cm} \\
& \textbf{Mean} & \textbf{41.1} & \textbf{59.9} & \textbf{54.0} & \textbf{51.7} & \textbf{43.9} & \textbf{49.9} & - &  & \\ \midrule
\multirow{6}{*}{\shortstack[l]{M3TR Expert \\ Models}} & $\map_{\overline{\mathrm{EL}}}$ & 51.7 & 69.4 & 56.3 & 55.4 & 49.7 & \textbf{56.5} & \textcolor{Green}{\textbf{+4.3}} & \multirow{7}{*}{$\mathcal{O}(\mathrm{n})$} & \multirow{7}{*}{\textcolor{Red}{\textbf{\xmark}}} \\
& $\map_{\overline{\mathrm{ER}}}$ & 44.8 & 66.5 & 57.0 & 57.8 & 48.7 & \textbf{55.0} & \textcolor{Green}{\textbf{+3.1}} &  & \\
& $\map_{\mathrm{BD}}$ & 40.2 & 57.3 & - & 54.7 & 49.2 & \textbf{50.2} & \textcolor{Green}{\textbf{+3.0}} &  & \\
& $\map_{\mathrm{CL}}$ & 45.1 & 63.2 & 61.1 & - & 48.6 & \textbf{55.0} & \textcolor{Green}{\textbf{+3.5}} &  & \\
& $\map_{\varnothing}$ & \textsuperscript{\textdagger}37.9 & \textsuperscript{\textdagger}55.0 & \textsuperscript{\textdagger}49.7 & \textsuperscript{\textdagger}48.2 & \textsuperscript{\textdagger}41.7 & \textsuperscript{\textdagger}\textbf{46.5} & \textcolor{Gray}{\textsuperscript{\textdagger}\textbf{+0.0}} &  & \\ \cdashline{2-9}
\vspace{-0.35cm} \\
& \textbf{Mean} & \textbf{43.9} & \textbf{62.3} & \textbf{56.0} & \textbf{54.0} & \textbf{47.5} & \textbf{52.6} & \textcolor{Green}{\textbf{+2.7}} &  & \\ \midrule
\multirow{6}{*}{\textbf{\shortstack[l]{M3TR\\Generalist}}} & $\map_{\overline{\mathrm{EL}}}$ & 48.8 & 67.8 & 59.5 & 54.8 & 51.8 & \textbf{56.5} & \textcolor{Green}{\textbf{+4.3}} & \multirow{7}{*}{$\mathcal{O}(1)$} & \multirow{7}{*}{\textcolor{Green}{\textbf{\cmark}}} \\
& $\map_{\overline{\mathrm{ER}}}$ & 45.7 & 64.4 & 57.0 & 56.9 & 51.1 & \textbf{55.0} & \textcolor{Green}{\textbf{+3.1}} &  & \\
& $\map_{\mathrm{BD}}$ & 41.2 & 57.3 & - & 53.0 & 48.0 & \textbf{49.9} & \textcolor{Green}{\textbf{+2.7}} &  & \\
& $\map_{\mathrm{CL}}$ & 42.5 & 59.3 & 57.4 & - & 45.6 & \textbf{51.2} & \textcolor{Gray}{\textbf{-0.3}} &  & \\
& $\map_{\varnothing}$ & 40.4 & 55.4 & 50.3 & 49.4 & 43.9 & \textbf{47.9} & \textcolor{Green}{\textbf{+1.4}} &  & \\ \cdashline{2-9}
\vspace{-0.35cm} \\ 
& \textbf{Mean} & \textbf{43.7} & \textbf{60.8} & \textbf{56.0} & \textbf{53.5} & \textbf{48.1} & \textbf{52.1} & \textcolor{Green}{\textbf{+2.2}} &  & \\ \bottomrule

\end{tabular}
}
\label{tab:eval_av2}
\end{table*}

\begin{table}
\vspace{0.35cm}
\centering
\caption{Results \emph{without} map masking as augmentation as ablation on the Argoverse 2 data set.}
\setlength\dashlinedash{1.2pt}
\setlength\dashlinegap{2.0pt}
\setlength\arrayrulewidth{0.3pt}
\resizebox{\linewidth}{!}{%
\begin{tabular}{wl{0.8cm} wr{0.7cm} wr{0.7cm} wr{0.7cm} wr{0.7cm} wr{0.7cm} wr{0.9cm} wr{0.9cm}}
\toprule
$\map_{p}$ & $\textbf{AP}^{\mathbfcal{C}}_{\textbf{dsh}}$ & $\textbf{AP}^{\mathbfcal{C}}_{\textbf{sol}}$ & $\textbf{AP}^{\mathbfcal{C}}_{\textbf{bou}}$ & $\textbf{AP}^{\mathbfcal{C}}_{\textbf{cen}}$ & $\textbf{AP}^{\mathbfcal{C}}_{\textbf{ped}}$ & $\textbf{mAP}^{\mathbfcal{C}}$ & \textbf{vs.~\cite{sun2024mapex}} \\ \toprule 
$\map_{\overline{\mathrm{EL}}}$ & 49.4 & 69.3 & 56.9 & 54.8 & 49.7 & \textbf{56.0} & \textcolor{Green}{\textbf{+3.8}} \\
$\map_{\overline{\mathrm{ER}}}$ & 41.4 & 65.6 & 55.8 & 56.2 & 48.6 & \textbf{54.4} & \textcolor{Green}{\textbf{+2.5}} \\
$\map_{\mathrm{BD}}$ & 42.7 & 58.4 & - & 52.7 & 46.4 & \textbf{49.7} & \textcolor{Green}{\textbf{+2.7}} \\
$\map_{\mathrm{CL}}$ & 42.7 & 59.9 & 55.4 & - & 43.6 & \textbf{50.4} & \textcolor{Gray}{\textbf{-1.1}} \\
$\map_{\varnothing}$ & 40.5 & 56.0 & 49.0 & 48.4 & 41.8 & \textbf{47.2} & \textcolor{Green}{\textbf{+0.7}} \\ \midrule 
%\cdashline{1-8}
\textbf{Mean} & \textbf{43.3} & \textbf{61.8} & \textbf{54.3} & \textbf{53.0} & \textbf{46.0} & \textbf{51.5} & \textcolor{Green}{\textbf{+1.6}} \\ \bottomrule

\end{tabular}
}
\label{tab:eval_gen_fix_prior}
\end{table}

\begin{table}
\vspace{0.35cm}
\centering
\caption{Comparison of map query encoders for the map prior scenario $\map_{\overline{\mathrm{EL}}}$ (ego lane is masked) on the Argoverse 2 dataset.}
\resizebox{\linewidth}{!}{%
\begin{tabular}{wl{1.2cm} wl{1.1cm} wc{1.4cm} wr{0.5cm} wr{0.5cm} wr{0.5cm} wr{0.5cm} wr{0.5cm}}
\toprule 
\multicolumn{2}{l}{\textbf{Map Query Enc.}} & & \multicolumn{5}{c}{$\textbf{AP}^{\mathbfcal{C}}$} \\ 
 Point Enc. & O2M\textsubscript{MMP} & $\textbf{mAP}^{\mathbfcal{C}}$ & dsh. & sol. & bou. & cen. & ped. \\ \toprule
A~\cite{sun2024mapex} & --- & 52.2 & 45.3 & 64.5 & 53.4 & 52.8 & 44.9 \\  
B & --- & 52.4 \textcolor{Green}{\textbf{(+0.2)}} & 46.8 & 65.3 & 53.0 & 52.5 & 44.5 \\
C & --- & 53.5 \textcolor{Green}{\textbf{(+1.3)}} & 48.4 & 66.7 & 55.4 & 50.6 & 46.5 \\
C & $\checkmark$ & 56.5 \textcolor{Green}{\textbf{(+4.3)}} & 51.7 & 69.4 & 56.3 & 55.4 & 49.7 \\ \bottomrule
\end{tabular}
}
\label{tab:map_query_enc_full}
\end{table}

%%%%%%%%%%%%%%%%%%%%%%%%%%%%%%%%%%%%%%%%%%%%%%%%%%%%%%%%%%%%%%%%%%%%%%%%%%%%%%%%%%%%%%%%%%%%

To simulate a real use case with various priors, groups of expert models are compared with a single generalist model, calculating the mean for each class across prior scenarios.
This assumes for the benefit of the experts that a perfect oracle for prior-to-model assignment exists and that no mixing of prior categories occurs.

%%%%%%%%%%%%%%%%%%%%%%%%%%%%%%%%%%%%%%%%%%%%%%%%%%%%%%%%%%%%%%%%%%%%%%%%%%%%%%%

\begin{table*}
\centering
\caption{
Comparison of methods and masking scenarios on the nuScenes data set, with the geographical split from~\cite{Lilja2024CVPR}.
Only elements not in the map prior are evaluated.
$\map_{\mathrm{BD}}$: Only road boundaries as prior.
$\map_{\mathrm{CL}}$: Only centerlines as prior.
$\map_{\varnothing}$: No map prior.
$\mathbfcal{O}(|\map_{\mathrm{P}}|)$ indicates how the deployment effort scales with the number of map prior types.
The last column indicates whether the method could handle variable priors without a not yet existing scenario to expert assignment oracle.
*: Re-implemented by the authors, as code was not publicly available at the time of publication.
\textdagger: For no map prior both expert methods are equivalent, as changes in the base architecture are only made regarding map priors.}
\setlength\dashlinedash{1.2pt}
\setlength\dashlinegap{2.0pt}
\setlength\arrayrulewidth{0.3pt}
\resizebox{0.95\linewidth}{!}{%
\begin{tabular}{l l wr{1.0cm} wr{1.0cm} wr{1.0cm} wr{1.0cm} wr{1.0cm} wr{1.0cm} wr{1.0cm} wc{1.2cm} wc{1.2cm}}
\toprule
\multicolumn{2}{c}{\small{\textit{Dataset: nuScenes}}} & \multicolumn{7}{c}{\small{\textit{$\textrm{AP}^{\mathcal{C}}$ = AP for Masked Elements only}}} & \multirow{2}{*}{\small{$\mathbfcal{O}(|\map_{\mathrm{P}}|)$}} & \multirow{2}{*}{\small{\textbf{\shortstack[c]{Var. Prior \\ w/o Oracle}}}} \\
\textbf{Method} & \textbf{Map Prior} & $\textbf{AP}^{\mathbfcal{C}}_{\textbf{dsh}}$ & $\textbf{AP}^{\mathbfcal{C}}_{\textbf{sol}}$ & $\textbf{AP}^{\mathbfcal{C}}_{\textbf{bou}}$ & $\textbf{AP}^{\mathbfcal{C}}_{\textbf{cen}}$ & $\textbf{AP}^{\mathbfcal{C}}_{\textbf{ped}}$ & $\textbf{mAP}^{\mathbfcal{C}}$ & \textbf{vs.~\cite{sun2024mapex}} & & \\ 
\vspace{-0.35cm} \\ \toprule 
MapTRv2~\cite{maptrv2} & $\map_{\varnothing}$ & 12.5 & 19.1 & 32.4 & 29.1 & 21.6 & \textbf{22.9} & \textcolor{Gray}{\textsuperscript{\textdagger}\textbf{+0.0}} & - & - \\ \midrule
\multirow{4}{*}{\shortstack[l]{MapEX*~\cite{sun2024mapex} \\ Models}} & $\map_{\mathrm{BD}}$ & 13.2 & 21.1 & - & 31.0 & 22.0 & \textbf{21.9} & - & \multirow{4}{*}{$\mathcal{O}(\mathrm{n})$} & \multirow{4}{*}{\textcolor{Red}{\textbf{\xmark}}} \\
& $\map_{\mathrm{CL}}$ & 16.6 & 26.0 & 39.8 & - & 23.4 & \textbf{26.4} & - &  & \\
& $\map_{\varnothing}$ & \textsuperscript{\textdagger}12.5 & \textsuperscript{\textdagger}19.1 & \textsuperscript{\textdagger}32.4 & \textsuperscript{\textdagger}29.1 & \textsuperscript{\textdagger}21.6 & \textsuperscript{\textdagger}\textbf{22.9} & - &  & \\ \cdashline{2-9}
\vspace{-0.35cm} \\
& \textbf{Mean} & \textbf{14.1} & \textbf{22.1} & \textbf{36.1} & \textbf{30.1} & \textbf{22.3} & \textbf{23.7} & - &  & \\ \midrule
\multirow{4}{*}{\shortstack[l]{M3TR Expert \\ Models}} & $\map_{\mathrm{BD}}$ & 15.3 & 26.7 & - & 34.9 & 28.3 & \textbf{26.3} & \textcolor{Green}{\textbf{+4.4}} & \multirow{4}{*}{$\mathcal{O}(\mathrm{n})$} & \multirow{4}{*}{\textcolor{Red}{\textbf{\xmark}}} \\
& $\map_{\mathrm{CL}}$ & 23.1 & 33.2 & 46.6 & - & 27.8 & \textbf{32.5} & \textcolor{Green}{\textbf{+6.1}} &  & \\
& $\map_{\varnothing}$ & \textsuperscript{\textdagger}12.5 & \textsuperscript{\textdagger}19.1 & \textsuperscript{\textdagger}32.4 & \textsuperscript{\textdagger}29.1 & \textsuperscript{\textdagger}21.6 & \textsuperscript{\textdagger}\textbf{22.9} & \textcolor{Gray}{\textsuperscript{\textdagger}\textbf{+0.0}} &  & \\ \cdashline{2-9}
\vspace{-0.35cm} \\
& \textbf{Mean} & \textbf{17.0} & \textbf{26.3} & \textbf{39.5} & \textbf{32.0} & \textbf{25.9} & \textbf{27.2} & \textcolor{Green}{\textbf{+3.5}} &  & \\ \midrule
\multirow{4}{*}{\textbf{\shortstack[l]{M3TR\\Generalist}}} & $\map_{\mathrm{BD}}$ & 14.5 & 23.2 & - & 32.7 & 24.7 & \textbf{23.8} & \textcolor{Green}{\textbf{+1.9}} & \multirow{4}{*}{$\mathcal{O}(1)$} & \multirow{4}{*}{\textcolor{Green}{\textbf{\cmark}}} \\
& $\map_{\mathrm{CL}}$ & 15.3 & 24.4 & 38.6 & - & 24.4 & \textbf{25.7} & \textcolor{Gray}{\textbf{-0.7}} &  & \\
& $\map_{\varnothing}$ & 12.4 & 20.0 & 31.8 & 29.8 & 23.4 & \textbf{23.5} & \textcolor{Green}{\textbf{+0.6}} &  & \\ \cdashline{2-9}
\vspace{-0.35cm} \\ 
& \textbf{Mean} & \textbf{14.1} & \textbf{22.5} & \textbf{35.2} & \textbf{31.3} & \textbf{24.2} & \textbf{24.3} & \textcolor{Green}{\textbf{+0.5}} &  & \\ \bottomrule

\end{tabular}
}
\label{tab:eval_nusc}
\end{table*}

%%%%%%%%%%%%%%%%%%%%%%%%%%%%%%%%%%%%%%%%%%%%%%%%%%%%%%%%%%%%%%%%%%%%%%%%%%%%%%%

\subsection{Implementation Details and Baseline}
\label{subsec:implementation}

We base our code and the model architecture on the MapTRv2~\cite{maptrv2} framework and re-implement MapEX~\cite{sun2024mapex} as a baseline.
Public code for~\cite{sun2024mapex} was not available at the time of writing and information about some of the query design particulars discussed in \cref{subsection:query_design} is not present in the paper. 
We therefore selected the variants \circled{A} and O2M\textsubscript{SMP} for the point query and query set design in our re-implementation.
All models use ResNet50~\cite{He_2016_CVPR} as the image backbone and parameters unrelated to map priors are left unchanged from the MapTRv2 base for fair comparison. 
We also follow one of the label modalities of MapTRv2 and use 3D map instances for Argoverse 2 as mentioned in \cref{tab:label_comparison}.

All models are trained until convergence, \ie for 24 / 110 epochs for experts and 54 / 224 epochs for the generalist on Argoverse~2 / nuScenes respectively, with the best checkpoint shown.
The generalist was trained on nine map prior scenarios for Argoverse 2 and seven for nuScenes.
The four scenarios not explicitly shown are missing only centerlines / pedestrian crossings / road borders / dividers. 

\subsection{Map Completion Performance}
\label{subsec:performance}

As we view \AV as our main dataset for evaluation, we investigate more map prior scenarios on it than on nuScenes. 
We first discuss the results on Argoverse 2 along with ablations on the map query encoder and the map masking as augmentation.
Then we present the slightly reduced set of experiments on the nuScenes dataset.

\subsubsection*{Results on Argoverse 2}

\begin{figure*}
    \centering
    % trim={<left> <lower> <right> <upper>}
    \includegraphics[trim={0 0.3cm 0 0},clip,width=\linewidth, center]{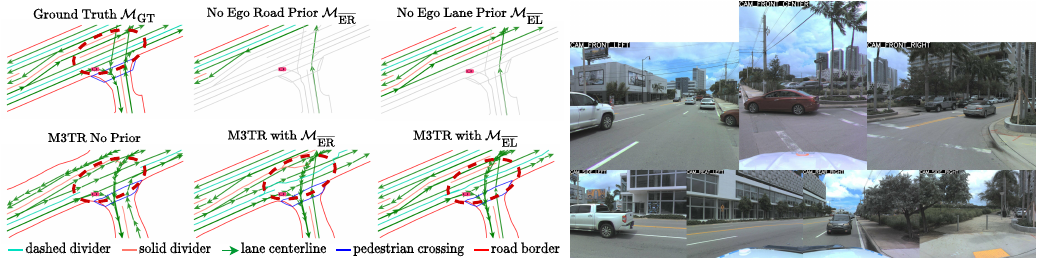}
    \caption{
    Example of the M3TR generalist model on the same sample from Argoverse 2 with different map priors.
    The more information available, the better the model can reconstruct elements not contained in the prior set.
    }
    \label{fig:qualitative_example}
\end{figure*}

\cref{tab:eval_av2} shows the performance of the M3TR expert and generalist variants as well as a MapEX expert baseline for five selected map prior scenarios on the Argoverse 2 dataset. 
All methods using map priors show enhanced average precision compared to the prior-less scenario, with varying benefit depending on the supplied map prior. 
A qualitative example of this varying benefit can be seen in \cref{fig:qualitative_example}.

For almost all scenarios, the M3TR experts variants substantially improve the prediction performance compared to the MapEX baseline.
Except for the $\map_{\mathrm{CL}}$ scenario, the generalist model matches the performance of the M3TR \expert models in their expert scenarios as well. 

In the more real-world use case with varying map priors, the generalist likewise outperforms the baseline models in the average of all scenarios, only using a fifth of the VRAM and without an oracle for perfect prior scenario to expert assignment. 
Such a scenario to expert assignment system does not exist yet and provides another substantial obstacle for real-world use of groups of expert models.

The generalist model additionally shows better performance in the no map prior scenario ($\mathcal{M}_{\varnothing}$), equivalent to the standard pure online HD map construction task, without architectural changes relevant for this scenario compared to the MapTRv2 base.
We conjecture that the various prior scenarios function as augmentation that also helps the model learn HD map construction without any prior.

\subsubsection*{Ablations on Argoverse 2}

The ablation in \cref{tab:eval_gen_fix_prior} highlights that using map masking as augmentation, \ie deriving each training data for each map prior scenario from the entire dataset is effective.
Compared to the naive prior generation (\cref{tab:eval_gen_fix_prior}), the generalist model with augmentation in \cref{tab:eval_av2} performs $+0.6$ mAP$^{\mathbfcal{C}}$ better.

The performance of the various modalities to encode map priors in queries can be seen in \cref{tab:map_query_enc_full}, using the point encoder names from \cref{fig:overview_flowchart}. 
Compared to the baseline encoder from MapEX~\cite{sun2024mapex}, \circled{A}, our proposed query design \circled{C} shows significantly improved performance. 
Encoder \circled{B}, which skips the modification of positional queries and reference points, has only a partial performance increase as a result.
Column $\mathrm{O2M_{MMP}}$ notes whether including the map prior in the one-to-many queries was enabled or not, leading to an even greater boost in performance.

\subsubsection*{Results on nuScenes}

Table \ref{tab:eval_nusc} shows results on the nuScenes dataset on a reduced set of map prior scenarios. 
Compared to the baseline, the increase in expert performance is even larger than on Argoverse 2.
While the \generalist improves upon the baseline, it shows reduced performance compared to M3TR experts.

Notably, the M3TR generalist model still shows increased performance for the case without prior compared to the MapTRv2 base, underlining once more the effectiveness of map masking as augmentation.

Along with the general decrease in \mapc compared to \AV, we hypothesize that the smaller sample count of the nuScenes dataset hampers generalization ability, thereby confirming an observation already noted in~\cite{Lilja2024CVPR}.

\section{Conclusion}
\label{sec:conclusion}

This work proposes M3TR, a novel generalist approach for HD map construction with variable map priors.

We introduce improved ground truth that moves the HD map construction task closer towards real HD planning maps.
Using it, we define a new HD map completion benchmark, including a systematic set of prior scenarios for outdated HD maps, combined with a metric that focuses on the elements not given as a map prior.

In terms of model design, we improve upon previous HD map construction models by systematically examining the query design to fully incorporate prior map information.
Experiments on \AV show that the proposed query point and query set design yields up to $+4.3$ \mapc compared to the MapEX~\cite{sun2024mapex} baseline.

We show that training with partially masked out maps not only allows using prior map information, but also serves as augmentation for HD map construction without any prior.

Finally, our novel \emph{Generalist} is a single model that can handle all map prior scenarios while matching the performance of specialized \experts.
Contrary to an ensemble of experts, it requires only constant memory and does not need to know which type of map information is available.
This makes M3TR the first real-world deployable model for HD map construction with offline HD map priors.

\section*{Acknowledgements}

We acknowledge the financial support by the German Federal Ministry of Education and Research (BMBF) within the project HAIBrid (FKZ 01IS21096D) and by the just better DATA (jbDATA) project supported by the German Federal Ministry for Economic Affairs and Climate Action of Germany (BMWK) and the European Union, grant number 19A23003H.
We thank our research partner Mercedes-Benz AG for the fruitful collaboration.
We also gratefully acknowledge financial support and computing resources provided by the Helmholtz Association’s Initiative and Networking Fund on HAICORE@FZJ.

% \clearpage

{
    %\newline
    %\todo{REFERENZEN CHECKEN!}
    \small
    \bibliographystyle{ieeenat_fullname}
    \bibliography{main}
}

% !!!! DELETE BEFORE SUBMISSION!!!!!
\clearpage
\setcounter{page}{1}
\maketitlesupplementary

\section{Remaining Details of Novel Ground Truth}
\label{sec:new_gt_supp}

This section presents more details of our novel ground truth, namely a more detailed description of the changes and their motivation, qualitative examples of commonly used and new labels and a comparison of label features on the nuScenes dataset.

\begin{figure}

    \centering
    \begin{subfigure}[t]{\linewidth}
        \centering
        \includegraphics[width=0.9\linewidth]{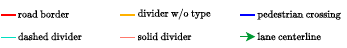} 
        %\label{fig:sub1}
    \end{subfigure}

    \centering
    \begin{subfigure}[t]{0.49\linewidth}
        \centering
        \includegraphics[width=\linewidth]{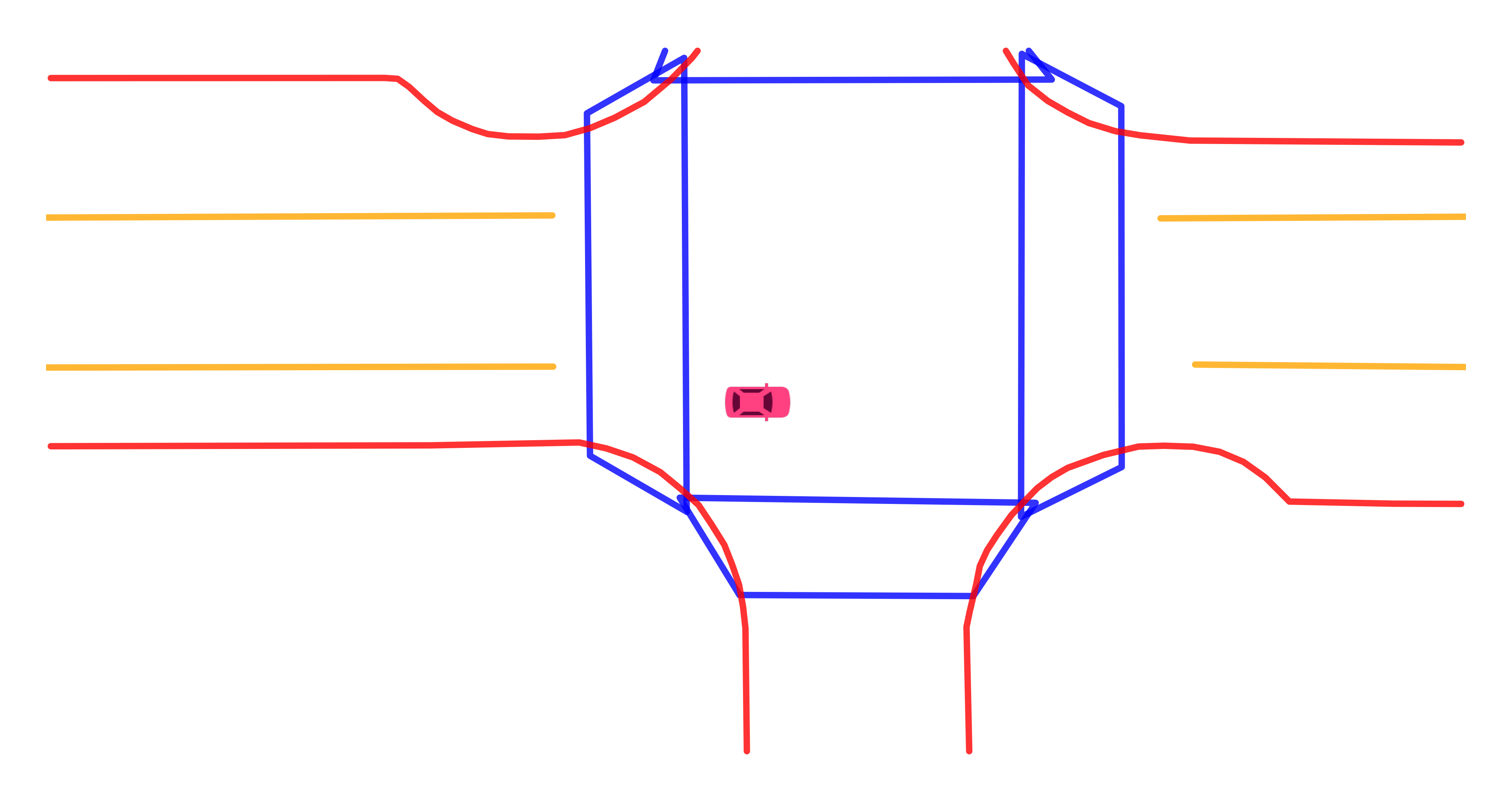} 
        %\label{fig:sub2}
    \end{subfigure}
    \begin{subfigure}[t]{0.49\linewidth}
        \centering
        \includegraphics[width=\linewidth]{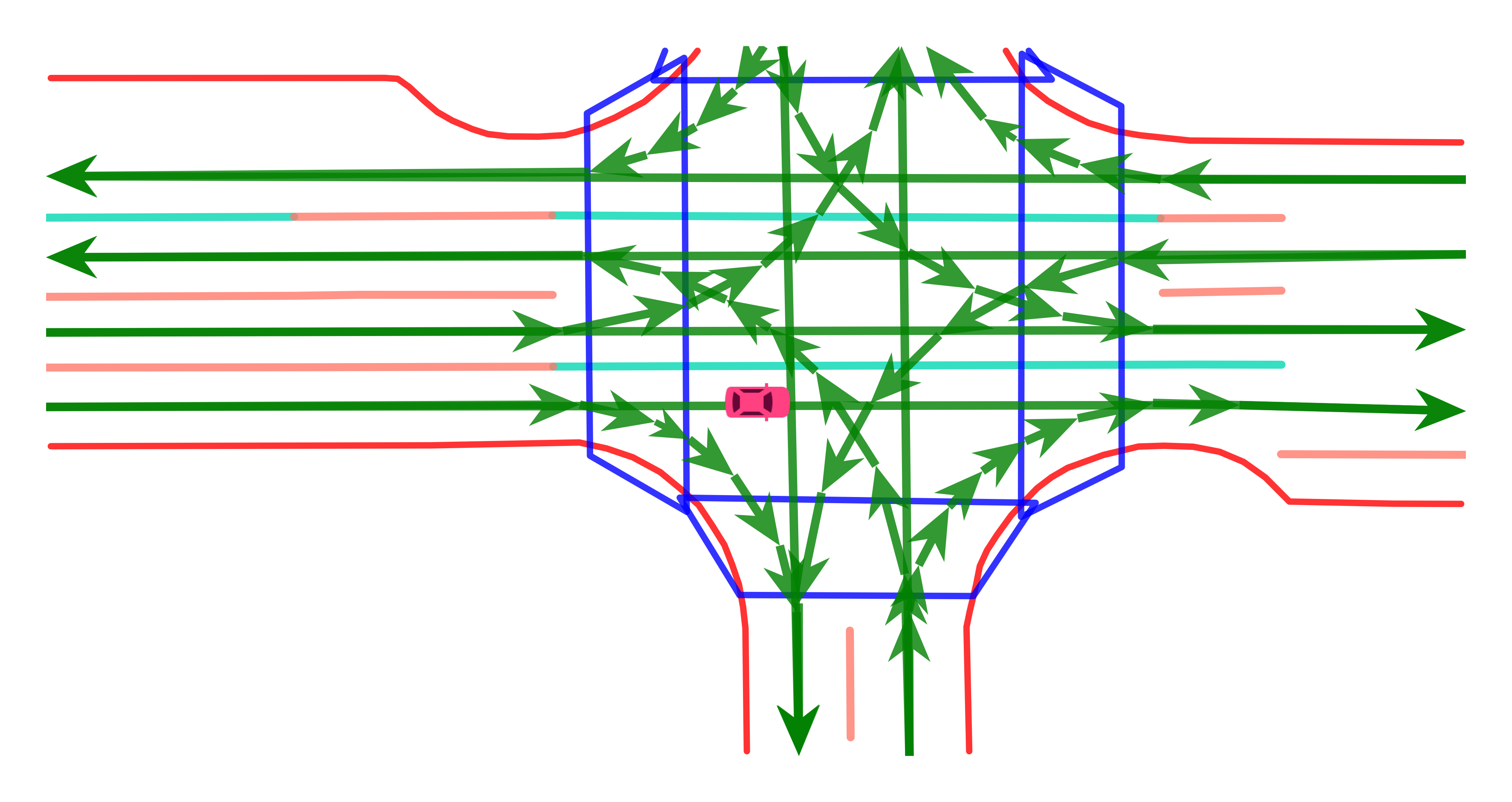} 
        %\label{fig:sub3}
    \end{subfigure}
    \vspace{-0.15cm}

    \begin{subfigure}[t]{0.49\linewidth}
        \centering
        \includegraphics[width=\linewidth]{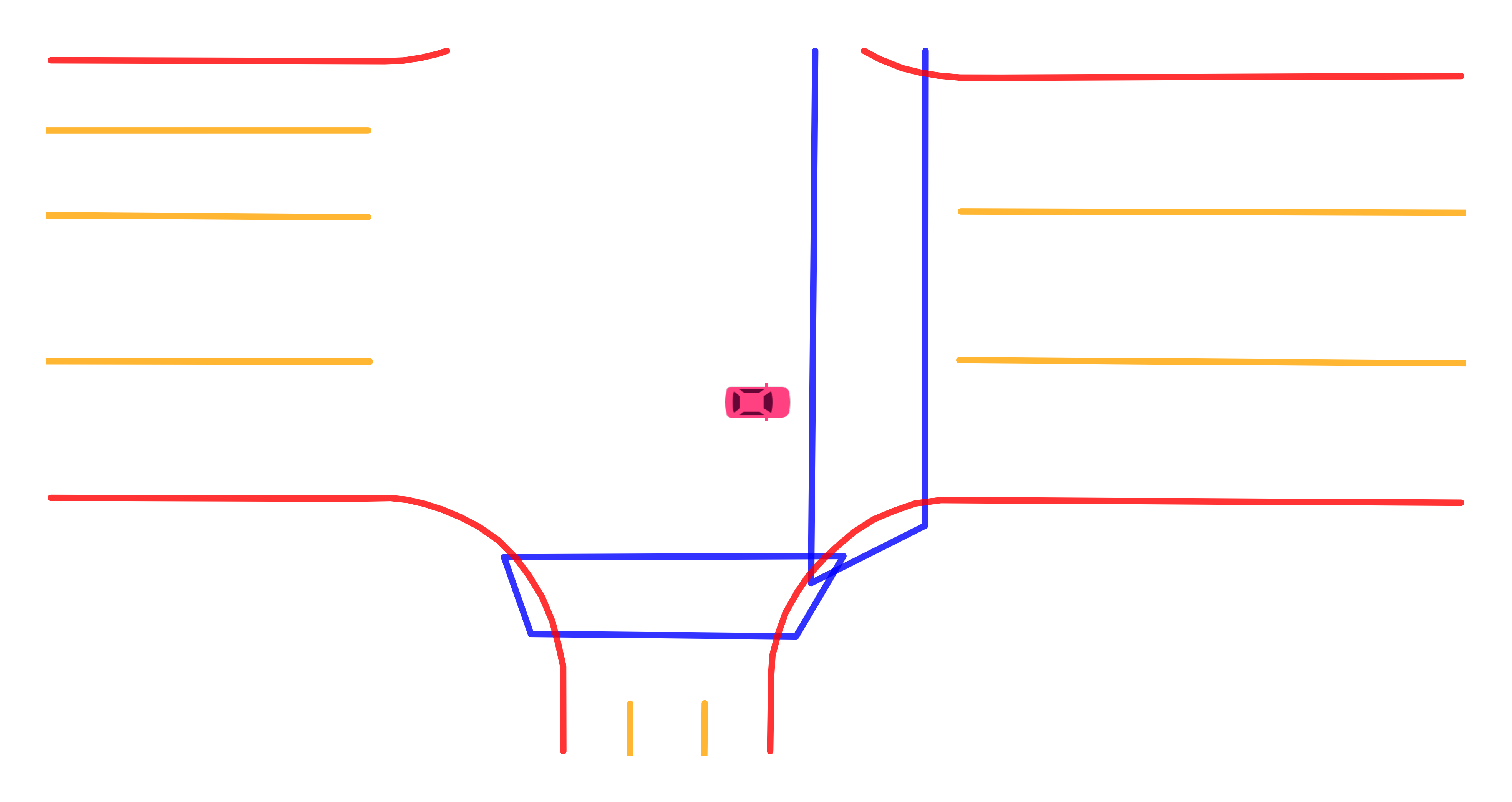} 
        %\label{fig:sub4}
    \end{subfigure}
    \begin{subfigure}[t]{0.49\linewidth}
        \centering
        \includegraphics[width=\linewidth]{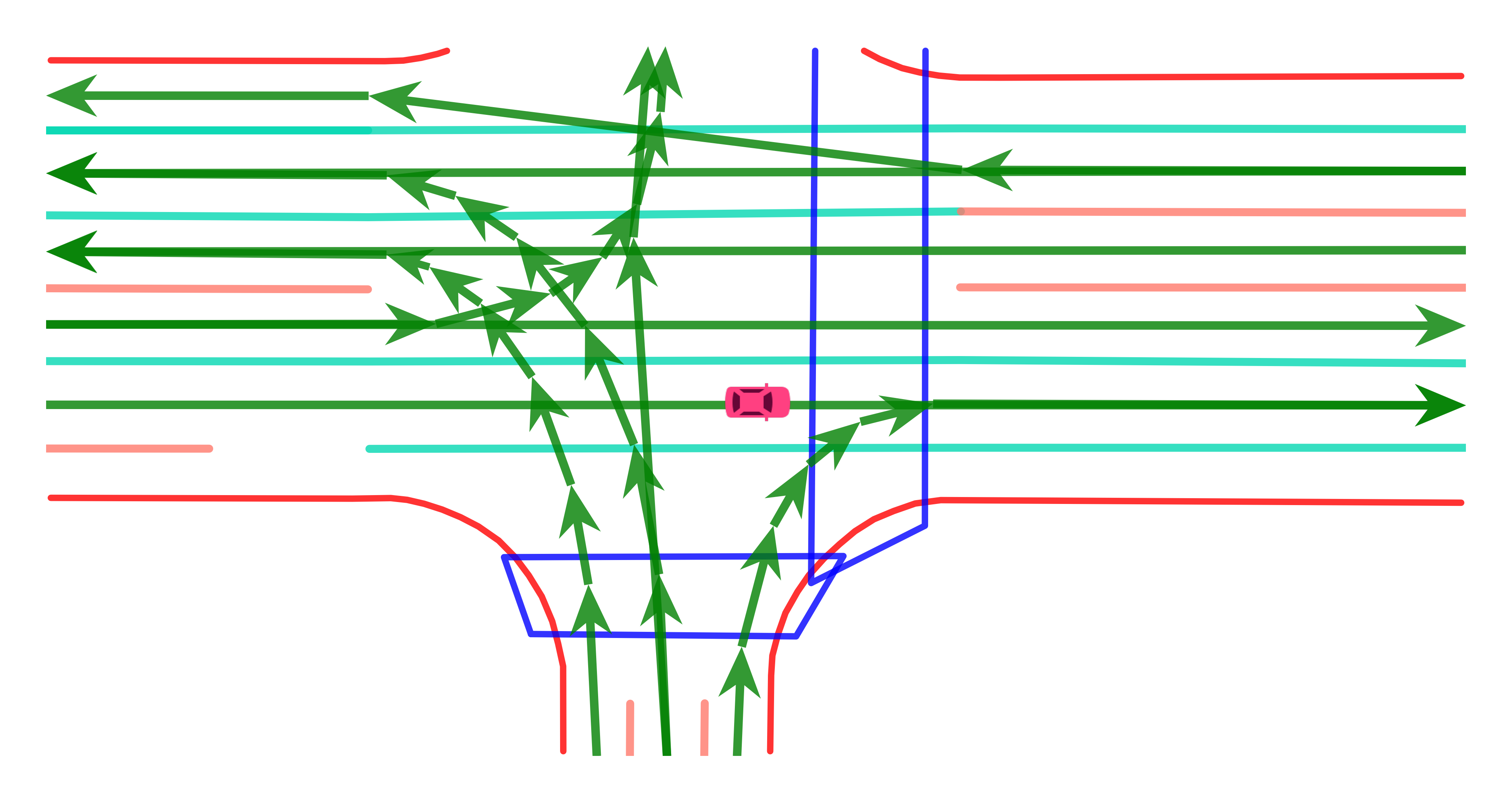} 
        %\label{fig:sub5}
    \end{subfigure}
    \vspace{-0.15cm}
    
    \begin{subfigure}[t]{0.49\linewidth}
        \centering
        \includegraphics[width=\linewidth]{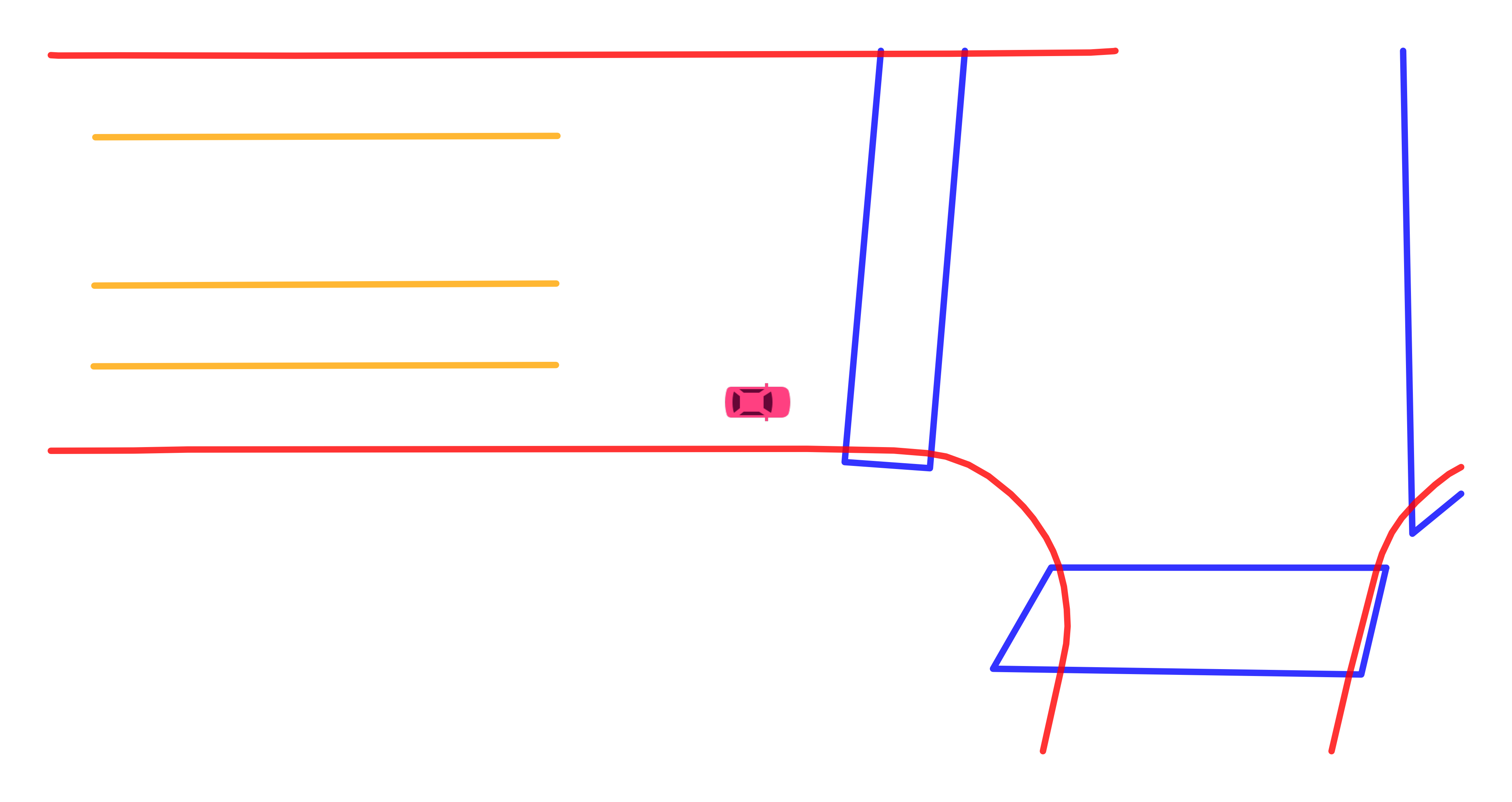} 
        \caption{Common labels \cite{maptr,Yuan_2024_streammapnet, zeng2024unifiedvectorpriorencoding, sun2024mapex}}
        %\label{fig:sub6}
    \end{subfigure}
    \begin{subfigure}[t]{0.49\linewidth}
        \centering
        \includegraphics[width=\linewidth]{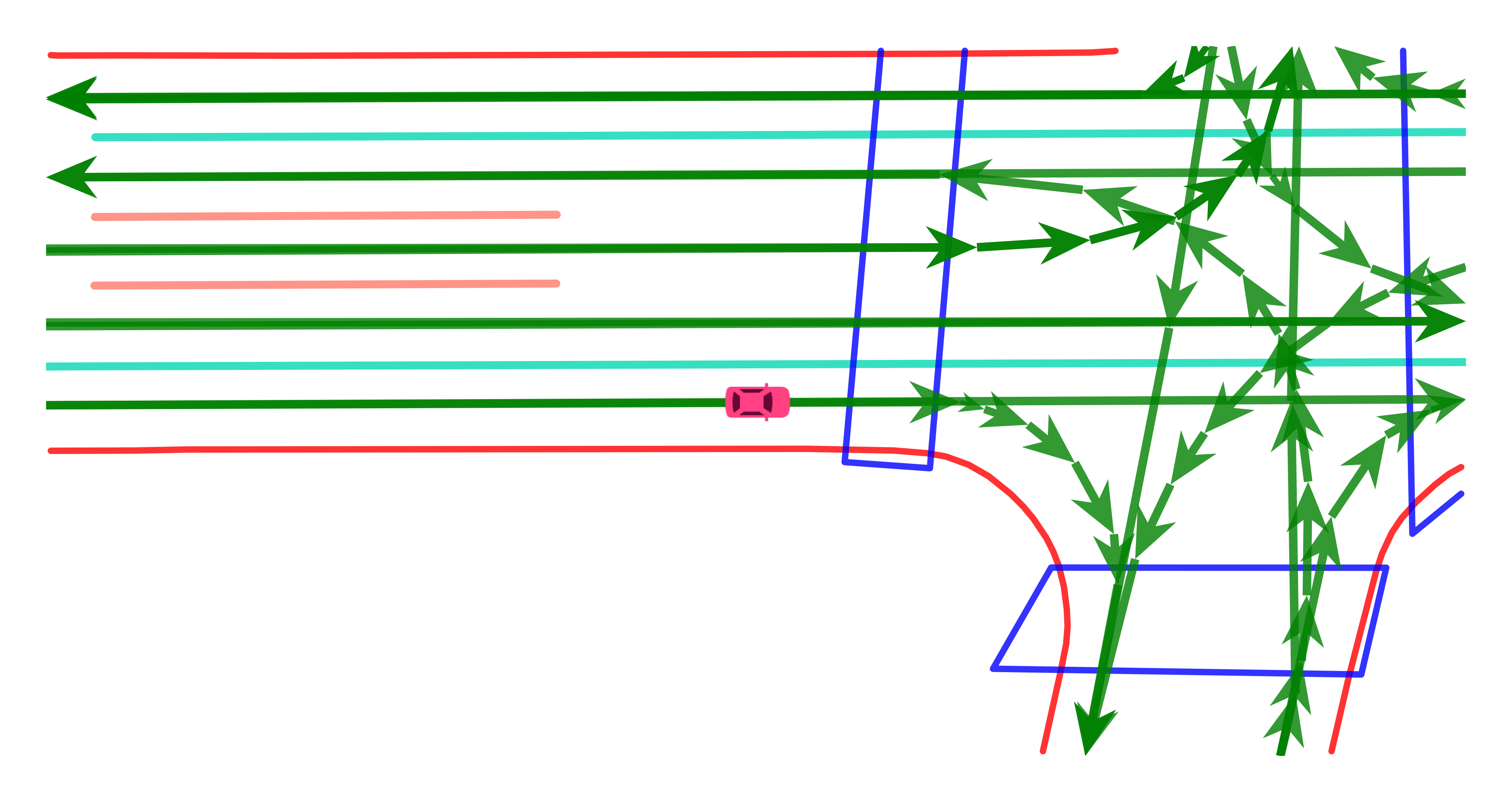} 
        \caption{Our proposed labels}
        %\label{fig:sub7}
    \end{subfigure}
    \caption{Comparisons of commonly used labels versus our proposed ground truth on Argoverse 2. }
    \label{fig:gt_comparison_supp}
\end{figure}

\subsubsection*{Detailed Description of Changes and Motivation}

Compared with semantic HD planning maps used in map-based autonomous driving stacks using \eg OpenDrive~\cite{ASAM_OpenDRIVE_2023} or Lanelet2~\cite{poggenhans2018lanelet2}, the shortcomings of commonly used labels become particularly evident. 
We describe these shortcomings, already mentioned in \cref{subsec:ground_truth_generation}, our changes and their motivation in more detail here.

Previously used labels distinguish only three classes of map elements: \emph{boundary}, \emph{divider} and \emph{pedestrian crossing}.
They hence lack information that is necessary for autonomous driving, such as the lane topology or the distinction between dashed and solid dividers which determines if lane change maneuvers are allowed. 
% Portability beyond existing public datasets is also very limited due to customized scripts and non-standardized map formats.
Recognizing this issue, LaneGAP~\cite{lanegap} introduces lane centerlines to represent the topology, but does it in a way that is compatible with other map elements.

Furthermore, as pointed out and also solved by~\cite{chen2024maptracker}, the label generation algorithms of \cite{vectormapnet,maptr,maptrv2,Yuan_2024_streammapnet,zeng2024unifiedvectorpriorencoding} produce missing or cut-off map element instances and merge actually independent elements inconsistently across frames.
The third issue with current labels is the geographic overlap of the data splits pointed out by~\cite{Lilja2024CVPR}, leading to leakage between training and evaluation data.

To solve all described issues, we provide new ground truth labels.
We complement the centerlines from \cite{lanegap,maptrv2}, which encode lane topology, with dashed and solid lane dividers as individual classes. 
%To solve the consistency issues, but also make the ground truth generation portable (\todo{WAS HEISST DAS GENAU?}), for Argoverse~2, we generate labels by converting the existing maps to the Lanelet2 framework~\cite{poggenhans2018lanelet2}.
For Argoverse~2, we generate labels by converting the original maps to Lanelet2~\cite{poggenhans2018lanelet2}, which provides not only a label generation algorithm~\cite{immel2024lanelet2mlconverter}, but also checks for the validity of ground truth maps. 
Similarly, in the future, Lanelet2 could serve as common label format across datasets, forming the basis for HD map construction foundation models.

\begin{table}[htb]
\caption{Comparison of labels on nuScenes used in various state of the art approaches with our proposed ground truth.}
\centering
\resizebox{0.8\columnwidth}{!}{
\begin{tabular}{l wc{1.0cm} wc{1.0cm} wc{1.0cm} } 
\multirow{2}{*}{\textbf{Method}} & \multirow{2}{*}{\shortstack[l]{Divider \\ Types}} & \multirow{2}{*}{\shortstack[l]{Lane \\ Centerl.}} & \multirow{2}{*}{\shortstack[l]{Geo. \\ Split}} \\ 
\\
\toprule
VectorMapNet \cite{vectormapnet} & - & - & - \\
MapTRv2 \cite{maptrv2} & - & ($\checkmark$) & - \\  
StreamMapNet \cite{Yuan_2024_streammapnet} & - & - & $\checkmark$ \\
MapTracker \cite{chen2024maptracker} & - & - & $\checkmark$ \\
MapEX \cite{sun2024mapex} & - & - & - \\
PriorDrive \cite{zeng2024unifiedvectorpriorencoding} & - & - & - \\
\textbf{M3TR (Ours)} & $\checkmark$ & $\checkmark$ & $\checkmark$ \\ 
\bottomrule
\end{tabular}
}
\label{tab:label_comparison_nusc}
\vspace{-1mm}
\end{table}

\subsubsection*{Qualitative Comparison}

\cref{fig:gt_comparison_supp} shows examples of commonly used labels and our ground truth. The new ground truth is semantically richer, distinguishes dashed and solid lane dividers and fixes the many missing or incorrect divider instances.

\begin{figure*}
    \centering
    \begin{subfigure}[t]{\linewidth}
        \centering
        \includegraphics[width=\linewidth]{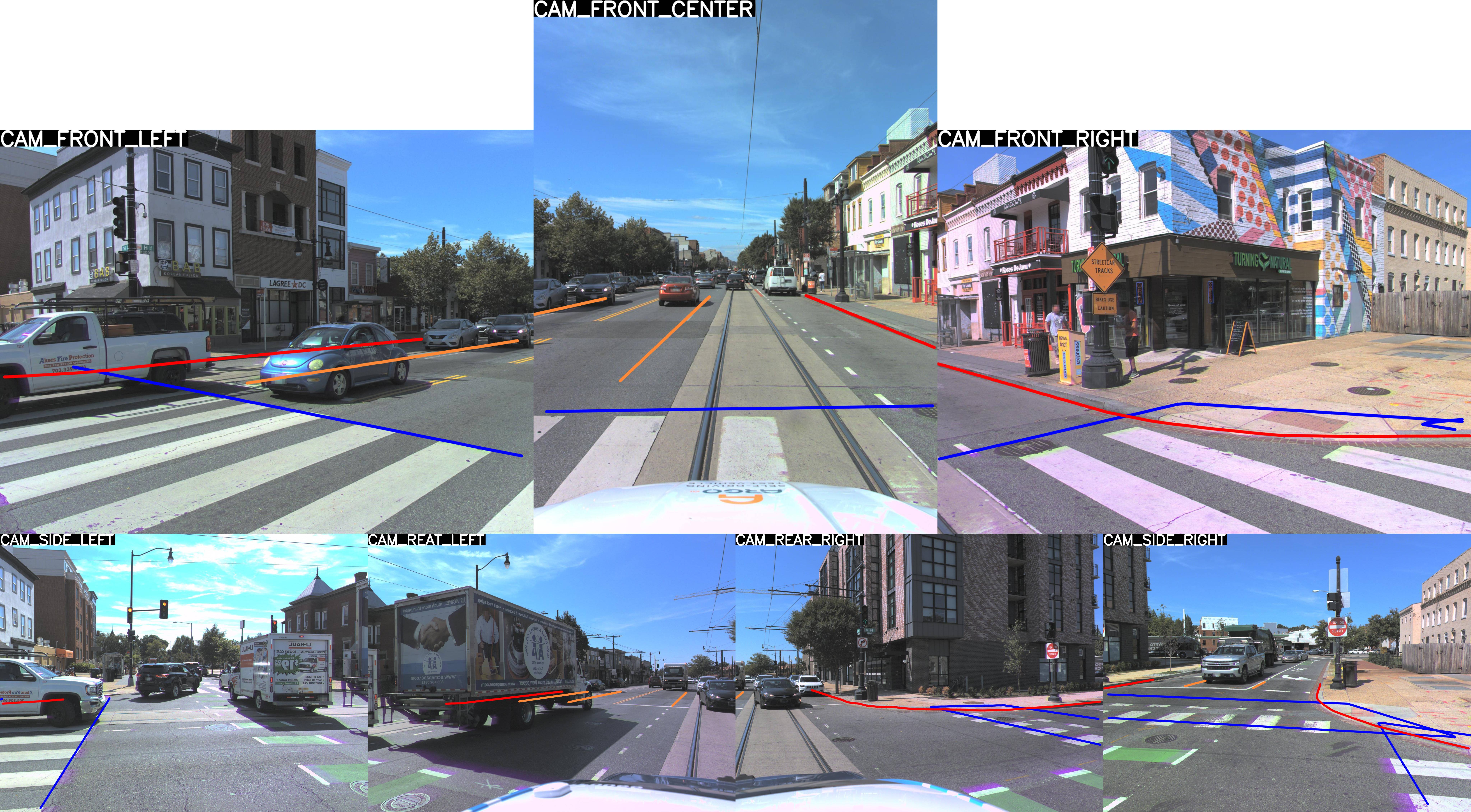} 
        \caption{Commonly used ground truth (generated from \cite{maptrv2}, as used in \cite{maptr, Yuan_2024_streammapnet, sun2024mapex, zeng2024unifiedvectorpriorencoding}).}
        \label{fig:sub1_label_projection}
    \end{subfigure}
    \hfill
    \begin{subfigure}[t]{\linewidth}
        \centering
        \includegraphics[width=\linewidth]{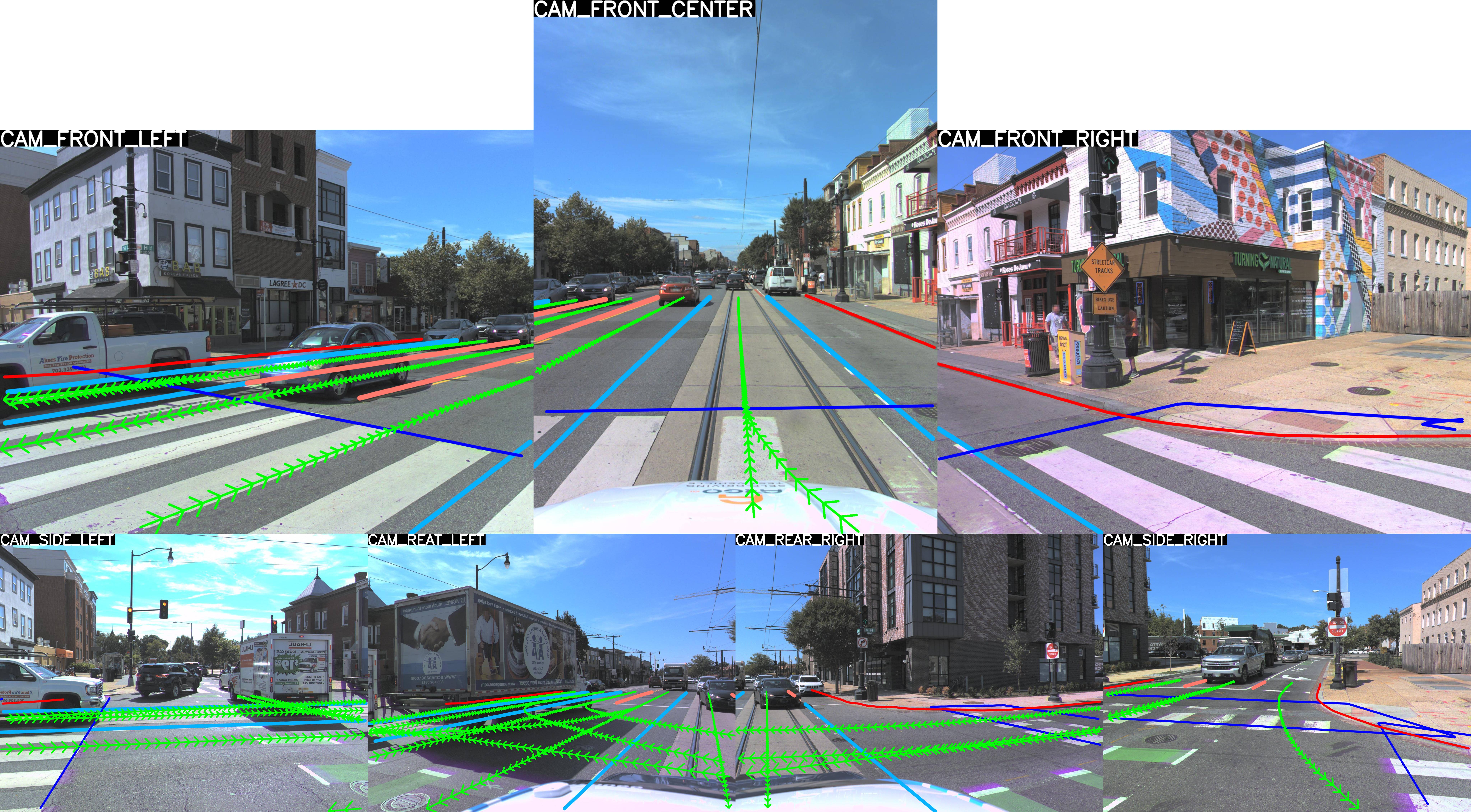} 
        \caption{Our new ground truth.}
        \label{fig:sub2_label_projection}
    \end{subfigure}
    \caption{Commonly used ground truth and our new ground truth on Argoverse 2 reprojected into the associated camera images. Many lane dividers are incorrect or missing from the common ground truth and semantically important distinctions between dashed and solid dividers are not present.}
    \label{fig:supp_label_projection}
\end{figure*}

\cref{fig:supp_label_projection} contrasts commonly used ground truth and our new ground truth for the same sample in Argoverse 2, reprojected into the camera images. \cref{fig:sub1_label_projection} with the commonly used ground truth illustrates that many lane dividers are not included in the labels, such as the left dashed divider of the vehicles ego lane and the solid divider in front of the vehicle separating the two driving directions. Additionally, the lack of distinction between solid and dashed dividers leaves crucial planning information for the vehicle missing.
\cref{fig:sub2_label_projection}, our proposed new ground truth, contains these features and, as seen in the reprojection, provides a much more comprehensive collection of road information.

\subsubsection*{Labels on nuScenes}

The nuScenes dataset has a different and less detailed map format compared to Argoverse 2 and as a result not all label features from \cref{tab:label_comparison} are applicable. \cref{tab:label_comparison_nusc} provides a similar overview for the nuScenes dataset. 
nuScenes does not include 3D map labels, making this aspect of comparison moot.
Furthermore, the original label generation code of VectorMapNet~\cite{vectormapnet} was designed for nuScenes, therefore the divider artifacts from Argoverse 2 are not present. 
It is important to note though that the original nuScenes maps \emph{themselves} have missing divider annotations, for example intersection areas do not have annotated lane dividers at all \cite{nuscenes}. 
Like for Argoverse 2, M3TR also separates lane dividers by type and uses lane centerlines for nuScenes.
The code to generate lane centerline labels was taken from MapTRv2~\cite{maptrv2}, as it already exists in the codebase, however no model trained with these labels is mentioned or evaluated in the MapTRv2 paper.

\section{Details of Loss Pre-Attribution}
\label{sec:loss_supp}

\begin{figure}
    \centering
    \includegraphics[width=\linewidth, center]{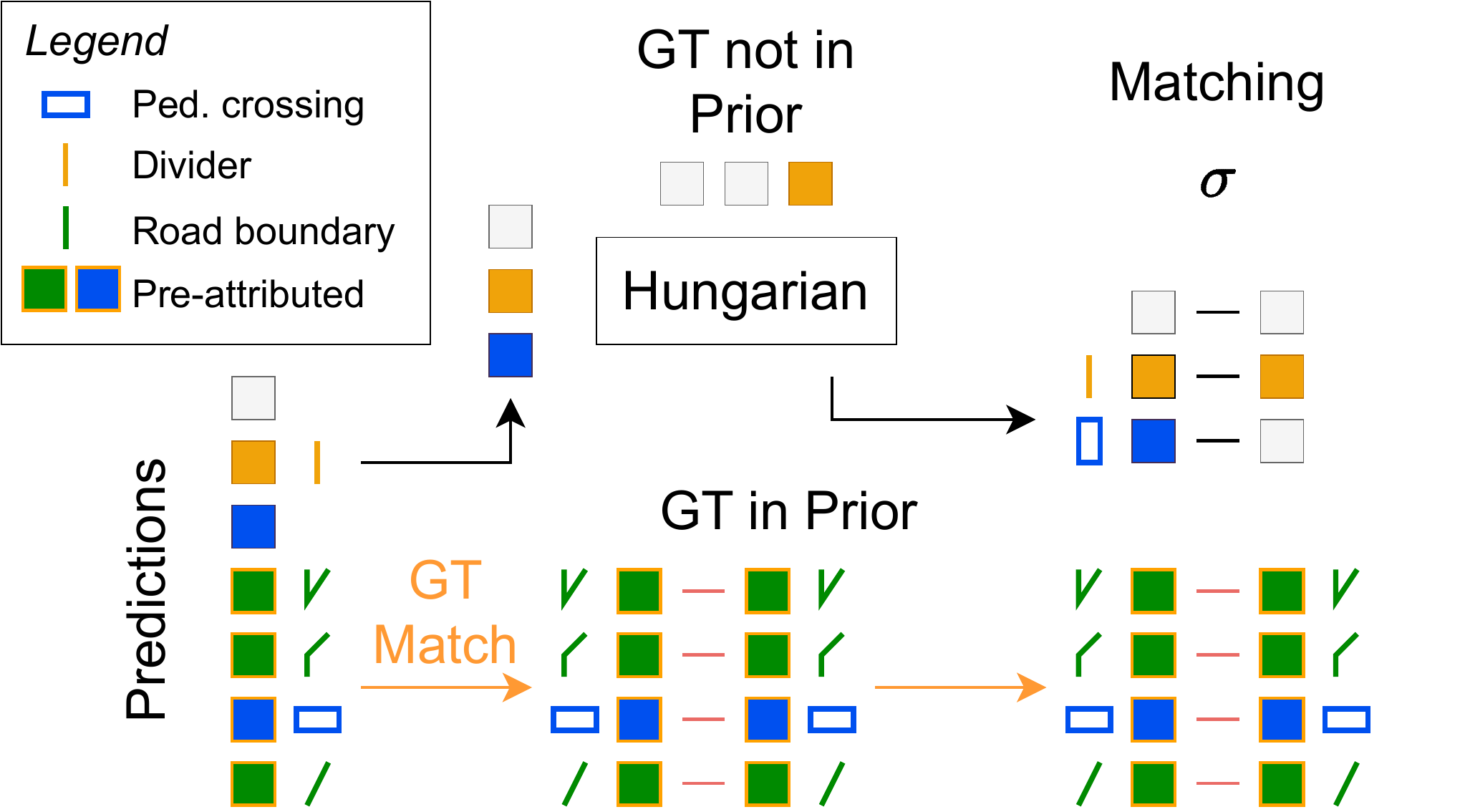}
    \caption{Visualization of the pre-attribution matching strategy from MapEX~\cite{sun2024mapex} (figure adapted from \cite{sun2024mapex}). To help the model in identifying map prior queries, ground truth instances in the prior are directly pre-assigned to the respective prediction before the Hungarian matching.}
    \label{fig:map_ex_matching}
\end{figure}

As mentioned in \cref{subsection:query_design}, we follow MapEX~\cite{sun2024mapex} for the training loss, including the pre-attribution of map prior instances during assignment. A visualization of the pre-attribution strategy can be seen in \cref{fig:map_ex_matching}.
All ground truth instances given to the model as a prior are also included in the loss term, so that the model is trained to also predict ground truth elements, receiving a complete map as output.
As a result, the model learns very quickly to pass through prior map elements for our task of map completion.

To help the model in identifying queries with map priors, MapEX proposes to remove instances of these queries from the regular Hungarian assignment and directly assign them to the respective ground truth elements, as it is already known for the initial map prior queries which ground truth instance they belong to.

\section{Additional Evaluation Results and Details}

%%%%%%%%%%%%%%%%%%%%%%%%%%%%%%%%%%%%%%%%%%%%%%%%%%%%%%%%%%%%%%%%%%%%%%%%%%%%%%%%%%%%%%%%%%%%

\begin{table*}
\centering
\caption{Comparison of methods over map prior scenarios on the Argoverse 2 data set, with the geographical split from~\cite{Lilja2024CVPR}.
\textbf{All} elements are evaluated, whether in the map prior or not.
*: Re-implemented by the authors, as code was not publicly available at the time of publication.
\textsuperscript{\textdagger}: AP values equal to $\textrm{AP}^{\mathcal{C}}$ as no elements of this class are in the prior.}
\setlength\dashlinedash{1.2pt}
\setlength\dashlinegap{2.0pt}
\setlength\arrayrulewidth{0.3pt}
\resizebox{0.75\linewidth}{!}{%
\begin{tabular}{l l wr{1.0cm} wr{1.0cm} wr{1.0cm} wr{1.0cm} wr{1.0cm} wr{1.0cm} wr{1.0cm}}
\toprule
\multicolumn{2}{c}{\small{\textit{Dataset: Argoverse 2}}} & \multicolumn{7}{c}{\small{\textit{$\textrm{AP}$ = AP for \textbf{All} Elements}}} \\
\textbf{Method} & \textbf{Map Prior} & $\textbf{AP}_{\textbf{dsh}}$ & $\textbf{AP}_{\textbf{sol}}$ & $\textbf{AP}_{\textbf{bou}}$ & $\textbf{AP}_{\textbf{cen}}$ & $\textbf{AP}_{\textbf{ped}}$ & $\textbf{mAP}$ & \textbf{vs.~\cite{sun2024mapex}}\\ 
\vspace{-0.35cm} \\ \toprule 
MapTRv2~\cite{maptrv2} & $\map_{\varnothing}$ & 37.9 & 55.0 & 49.7 & 48.2 & 41.7 & \textbf{46.5} & - \\ \midrule
\multirow{6}{*}{\shortstack[l]{MapEX*~\cite{sun2024mapex} \\ Models}} & $\map_{\overline{\mathrm{EL}}}$ & 77.4 & 85.5 & 97.1 & 79.3 & 58.9 & \textbf{79.7} & - \\
& $\map_{\overline{\mathrm{ER}}}$ & 67.3 & 74.6 & 94.7 & 70.5 & 54.4 & \textbf{72.3} & - \\
& $\map_{\mathrm{BD}}$ & \textsuperscript{\textdagger}37.7 & \textsuperscript{\textdagger}56.0 & 97.1 & \textsuperscript{\textdagger}50.6 & \textsuperscript{\textdagger}44.5 & \textbf{57.2} & - \\
& $\map_{\mathrm{CL}}$ & \textsuperscript{\textdagger}43.2 & \textsuperscript{\textdagger}61.8 & \textsuperscript{\textdagger}58.1 & 95.7 & \textsuperscript{\textdagger}42.8 & \textbf{60.3} & - \\ \cdashline{2-9}
\vspace{-0.35cm} \\
& \textbf{Mean} & \textbf{56.4} & \textbf{69.5} & \textbf{86.8} & \textbf{74.0} & \textbf{50.2} & \textbf{67.4} & -  \\ \midrule
\multirow{6}{*}{\shortstack[l]{M3TR Expert \\ Models}} & $\map_{\overline{\mathrm{EL}}}$ & 80.5 & 87.6 & 97.1 & 80.6 & 61.0 & \textbf{81.4} & \textcolor{Green}{\textbf{+1.7}} \\
& $\map_{\overline{\mathrm{ER}}}$ & 69.4 & 77.2 & 94.7 & 72.7 & 55.4 & \textbf{73.9} & \textcolor{Green}{\textbf{+1.6}} \\
& $\map_{\mathrm{BD}}$ & \textsuperscript{\textdagger}40.2 & \textsuperscript{\textdagger}57.3 & 99.9 & \textsuperscript{\textdagger}54.7 & \textsuperscript{\textdagger}49.2 & \textbf{60.3} & \textcolor{Green}{\textbf{+3.1}} \\
& $\map_{\mathrm{CL}}$ & \textsuperscript{\textdagger}45.1 & \textsuperscript{\textdagger}63.2 & \textsuperscript{\textdagger}61.1 & 96.3 & \textsuperscript{\textdagger}48.6 & \textbf{62.9} & \textcolor{Green}{\textbf{+2.6}} \\  \cdashline{2-9}
\vspace{-0.35cm} \\
& \textbf{Mean} & \textbf{58.8} & \textbf{71.3} & \textbf{88.2} & \textbf{76.1} & \textbf{53.6} & \textbf{69.6} & \textcolor{Green}{\textbf{+2.3}} \\ \midrule
\multirow{6}{*}{\textbf{\shortstack[l]{M3TR\\Generalist}}} & $\map_{\overline{\mathrm{EL}}}$ & 78.9 & 86.8 & 97.6 & 80.3 & 63.3 & \textbf{81.4} & \textcolor{Green}{\textbf{+1.7}} \\
& $\map_{\overline{\mathrm{ER}}}$ & 69.5 & 75.8 & 95.0 & 72.6 & 57.7 & \textbf{74.1} & \textcolor{Green}{\textbf{+1.8}} \\
& $\map_{\mathrm{BD}}$ & \textsuperscript{\textdagger}41.2 & \textsuperscript{\textdagger}57.3 & 99.9 & \textsuperscript{\textdagger}53.0 & \textsuperscript{\textdagger}48.0 & \textbf{59.9} & \textcolor{Green}{\textbf{+2.7}} \\
& $\map_{\mathrm{CL}}$ & \textsuperscript{\textdagger}42.5 & \textsuperscript{\textdagger}59.3 & \textsuperscript{\textdagger}57.4 & 96.9 & \textsuperscript{\textdagger}45.6 & \textbf{60.3} & \textcolor{Gray}{\textbf{+0.0}} \\ \cdashline{2-9}
\vspace{-0.35cm} \\ 
& \textbf{Mean} & \textbf{58.0} & \textbf{69.8} & \textbf{87.5} & \textbf{75.7} & \textbf{53.7} & \textbf{68.9} & \textcolor{Green}{\textbf{+1.6}} \\ \bottomrule

\end{tabular}
}
\label{tab:eval_av2_all_elements}
\end{table*}

%%%%%%%%%%%%%%%%%%%%%%%%%%%%%%%%%%%%%%%%%%%%%%%%%%%%%%%%%%%%%%%%%%%%%%%%%%%%%%%%%%%%%%%%%%%%

This section presents additional evaluation results on Argoverse 2 and qualitative examples to complement \cref{sec:experiments}.

\subsubsection*{Remaining Implementation Details}

As shown in \cref{fig:o2m_queries}, the total number of O2M queries is chosen as a multiple of the ground truth repetitions to match with the tiling. In total all models have $70$ O2O and $350$ O2M queries, therefore having $5$ times the amount of O2O queries as O2M queries. MapTRv2~\cite{maptrv2} uses a fixed number of 300 O2M queries in their original implementation.

\subsubsection*{Argoverse 2 AP with Prior}

For Argoverse 2, we also report the average precision for \emph{all} elements in the ground truth $\map_{\mathrm{GT}}$ in \cref{tab:eval_av2_all_elements}, whether they were included in the prior $\map_{\mathrm{P}}$ or not. 

In conjunction with the $\textrm{AP}^{\mathcal{C}}$ from \cref{tab:eval_av2}, this can give an overview on the total accuracy of the resulting map from the model depending on the supplied prior.
For elements in the prior, the AP is calculated with the output of the model for those elements, not using the original ground truth prior. 
Additionally, we can also evaluate the ability to pass through given prior elements. 
In an application setting, one would in any case include known map elements in the constructed HD map from the prior directly to have these elements be guaranteed correct.

All models reproduce the prior almost perfectly, with M3TR showing slightly better metrics for the $\map_{\mathrm{BD}}$ and $\map_{\mathrm{CL}}$ priors.
The M3TR \experts and \generalist also still outperform the baseline in regular AP, mirroring \cref{sec:experiments}, though less pronounced for the $\map_{\overline{\mathrm{EL}}}$ and $\map_{\overline{\mathrm{ER}}}$ scenarios.
This can be attributed to the fact that the prior, which both methods pass through very well, is now included in the metric. 
The share of online perceived instances, where the performance differences lie, is therefore diluted in the regular AP evaluated here.

\subsubsection*{Qualitative Results}

In the following, we show qualitative results of M3TR together with MapTRv2~\cite{maptrv2} and MapEX~\cite{sun2024mapex} on the Argoverse~2~\cite{Argoverse2} and nuScenes~\cite{nuscenes} datasets.

\begin{figure*}
    \centering
    \includegraphics[width=\linewidth]{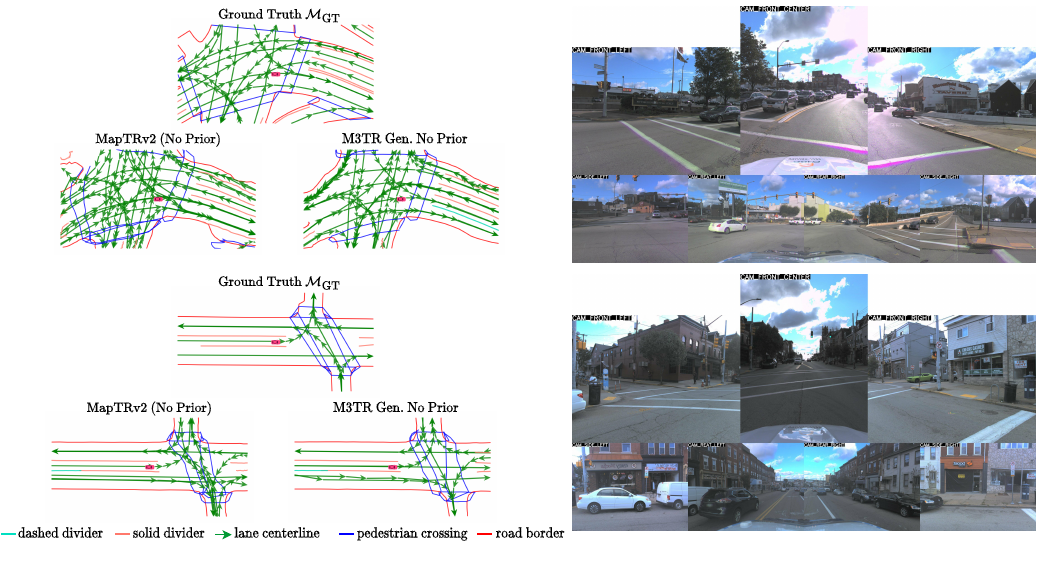} 
    \caption{Qualitative examples comparing the M3TR \generalist without any prior with MapTRv2~\cite{maptrv2} on Argoverse~2. M3TR shows improved performance even without any prior due to our proposed masking as augmentation.}
    \label{fig:qual_av2_1}
\end{figure*}

\cref{fig:qual_av2_1} displays the M3TR \generalist and MapTRv2 on Argoverse~2, both with no
prior. With our masking as augmentation training regime M3TR shows increased performance in complex scenes, without any architectural changes.

\begin{figure*}
    \centering
    \includegraphics[width=\linewidth]{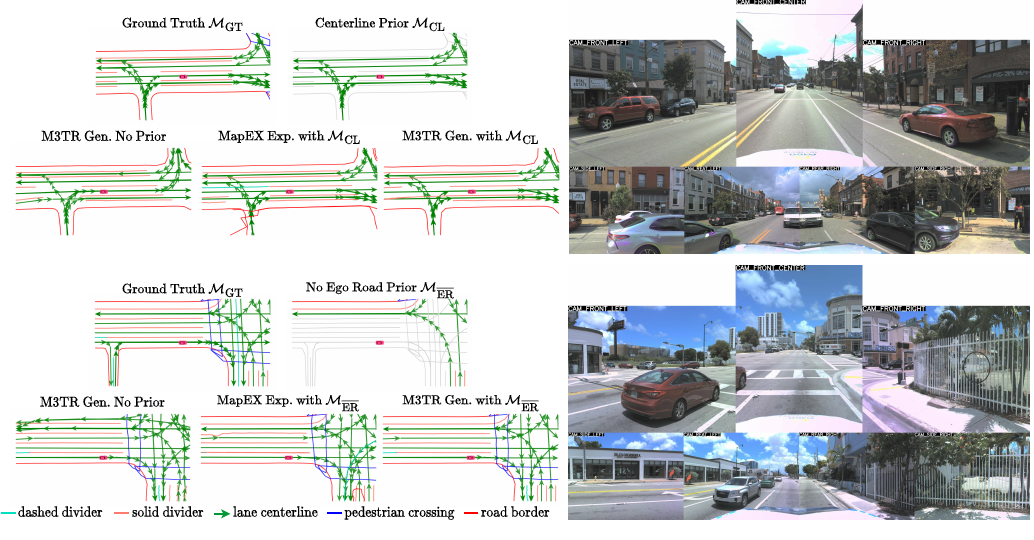} 
    \caption{Qualitative examples comparing the M3TR \generalist with the respective MapEX~\cite{sun2024mapex} \experts on Argoverse~2. }
    \label{fig:qual_av2_2}
\end{figure*}

\begin{figure*}
    \centering
    \includegraphics[width=\linewidth]{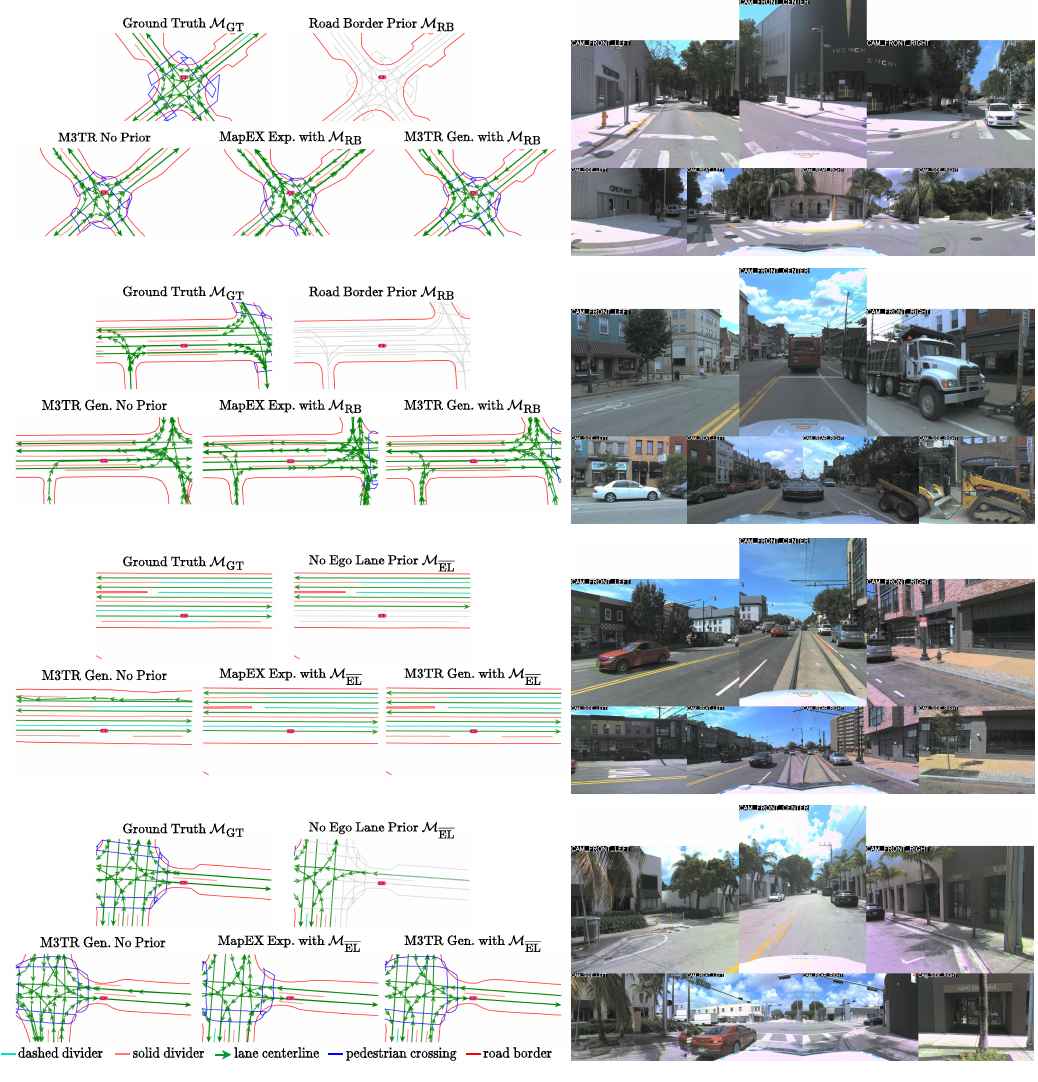} 
    \caption{More qualitative examples comparing the M3TR \generalist with the respective MapEX~\cite{sun2024mapex} \experts on Argoverse~2. The M3TR \generalist is able to perceive elements online with higher accuracy across various prior scenarios, while only requiring a single model.}
    \label{fig:qual_av2_3}
\end{figure*}

\cref{fig:qual_av2_2} and \cref{fig:qual_av2_3} demonstrate the M3TR \generalist and the respective MapEX \experts on every evaluated prior scenario and show the prediction of M3TR with no prior as well. The M3TR \generalist is able to use the prior to better perceive the missing online elements, with higher accuracy compared to the MapEX baseline and only using a single model. As seen in \cref{fig:qual_av2_2}, we further observed that MapEX sometimes inserts predictions that are logically inconsistent with the given prior, e.g. road borders that cross a lane centerline, hinting at a lessened understanding of the prior compared to M3TR. 

\begin{figure*}
    \centering
    \includegraphics[width=\linewidth]{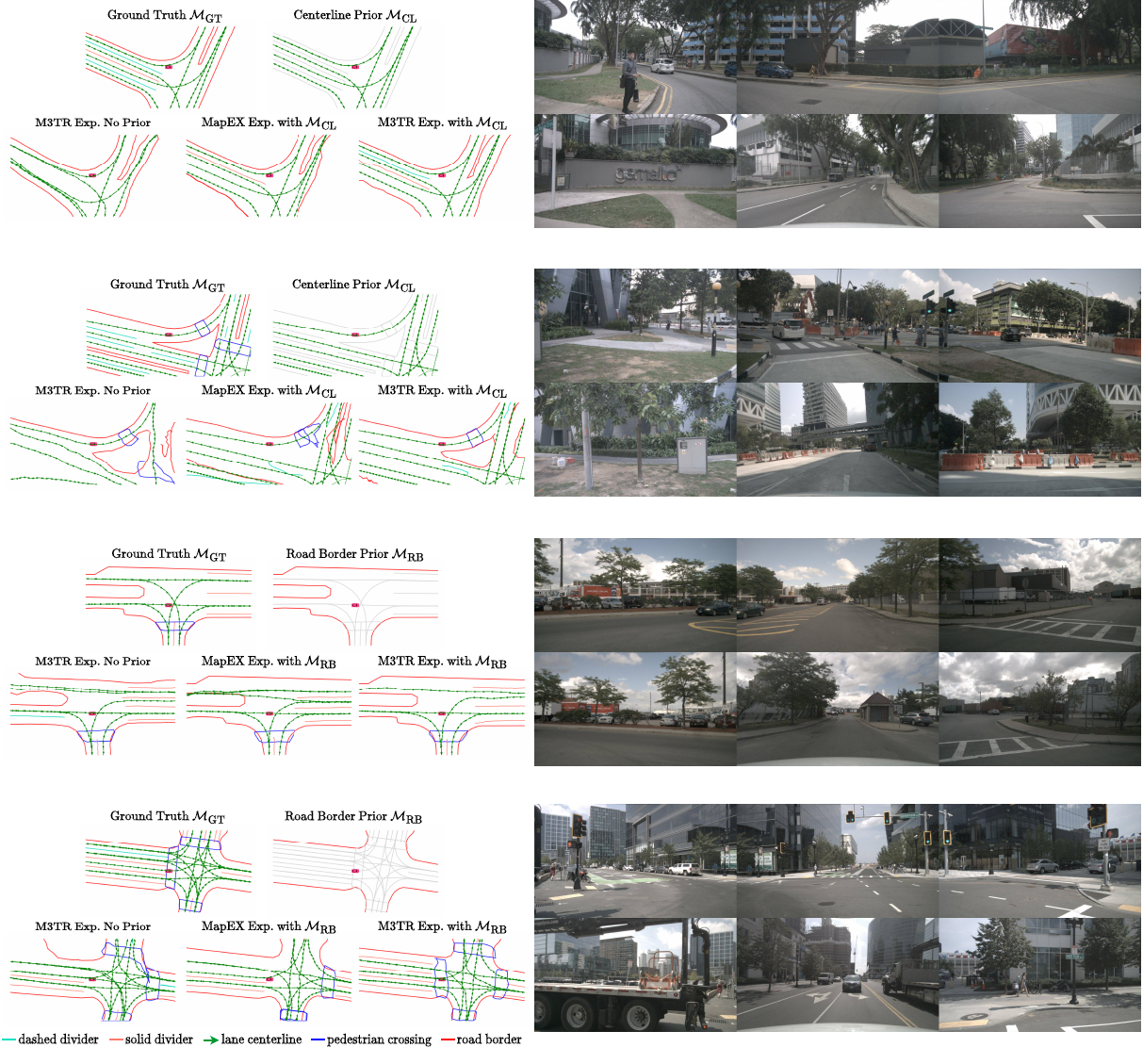} 
    \caption{Qualitative examples comparing the M3TR \experts with the respective MapEX~\cite{sun2024mapex} \experts on nuScenes.}
    \label{fig:qual_nusc}
\end{figure*}

For nuScenes, \cref{fig:qual_nusc} presents a comparison of the M3TR and MapEX \experts, with the no prior expert also shown for reference. 
The general performance is decreased of all models compared to Argoverse~2 with large errors without any prior. 
In conjunction, the performance gain when supplying a prior is stronger. 
The mentioned tendency of MapEX to sometimes make predictions that are logically inconsistent with the given prior can also be seen here.

\end{document}